\documentclass[a4paper,11pt]{article}
\usepackage{graphicx} 
\usepackage{amsmath}
\usepackage{float}
\usepackage[utf8]{inputenc}
\usepackage{wrapfig}
\usepackage{booktabs}
\usepackage{makecell}
\usepackage{array}
\usepackage{comment}
\usepackage{adjustbox}
\usepackage{hyperref}
\graphicspath{ {./images/} }
\usepackage{caption}
\usepackage{authblk}
\newcommand\myfrac[2]{\frac{#1}{#2\mathstrut}}

\usepackage[a4paper, total={6.5in, 8in}]{geometry}

\renewcommand{\arraystretch}{1.5}
\title{Active InSAR monitoring of building damage in Gaza during the Israel-Hamas War}

\author[1, 2]{Corey Scher*}
\author[2]{Jamon Van Den Hoek}
\affil[1]{Department of Earth and Environmental Sciences, City University of New York Graduate Center}
\affil[2]{Geography Program, College of Earth, Ocean and Atmospheric Sciences, Oregon State University}
\date{}

\begin{document}

\maketitle

\centerline{\textbf{Abstract}}
\vspace{0.5cm}

Aerial bombardment of the Gaza Strip beginning October 7, 2023 is one of the most intense bombing campaigns of the twenty-first century, driving widespread urban damage. Characterizing damage over a geographically dynamic and protracted armed conflict requires active monitoring. Synthetic aperture radar (SAR) has precedence for mapping disaster-induced damage with bi-temporal methods but applications to active monitoring during sustained crises are limited. Using interferometric SAR data from Sentinel-1, we apply a long temporal-arc coherent change detection (LT-CCD) approach to track weekly damage trends over the first year of the 2023- Israel-Hamas War. We detect 92.5\% of damage labels in reference data from the United Nations with a negligible (1.2\%) false positive rate. The temporal fidelity of our approach reveals rapidly increasing damage during the first three months of the war focused in northern Gaza, a notable pause in damage during a temporary ceasefire, and surges of new damage as conflict hot-spots shift from north to south. Three-fifths (191,263) of all buildings are damaged or destroyed by the end of the study. With massive need for timely data on damage in armed conflict zones, our low-cost and low-latency approach enables rapid uptake of damage information at humanitarian and journalistic organizations.


\section{Introduction}


The 2023- Israel-Hamas War (the war) began following a surprise attack on Israeli military targets and civilians by Palestinian militant groups operating from within the Gaza Strip on October 7, 2023. Over 1,200 Israeli civilians, military personnel, and internationals were killed and another 254 Israelis and internationals were taken hostage \cite{oct7_attacks}. That day, Israel began an aerial warfare campaign in Gaza followed by ground invasions, resulting in widespread urban destruction and the deaths of over 55,000 Palestinians as of writing \cite{Masquelier-Page2025Jun}. The aerial warfare campaign in Gaza has been described by military historians as one of the most intense conventional bombing campaigns since WWII \cite{Hi2023Dec}. Widespread urban destruction resulting from aerial bombardment has direct impacts on civilians, causing conflict-induced population displacement, affecting routes for safe passage, challenging the ability for humanitarian organizations to conduct recovery and relief work, and inducing long-term impacts to environment, public health, and economies \cite{VanDenHoek2021Sep}. Tracking armed conflict-induced damage is integral to support decision-making for humanitarians responding to conflict events and enable broader public understanding of conflict impacts while wars play out. The dynamic, fast-paced, and reported high-intensity of damage across Gaza is difficult to characterize, requiring a low-latency and sustained mapping of damage across the duration of the ongoing war.

\begin{figure}[H]
    \centering
    \centerline{\includegraphics[width=20cm]{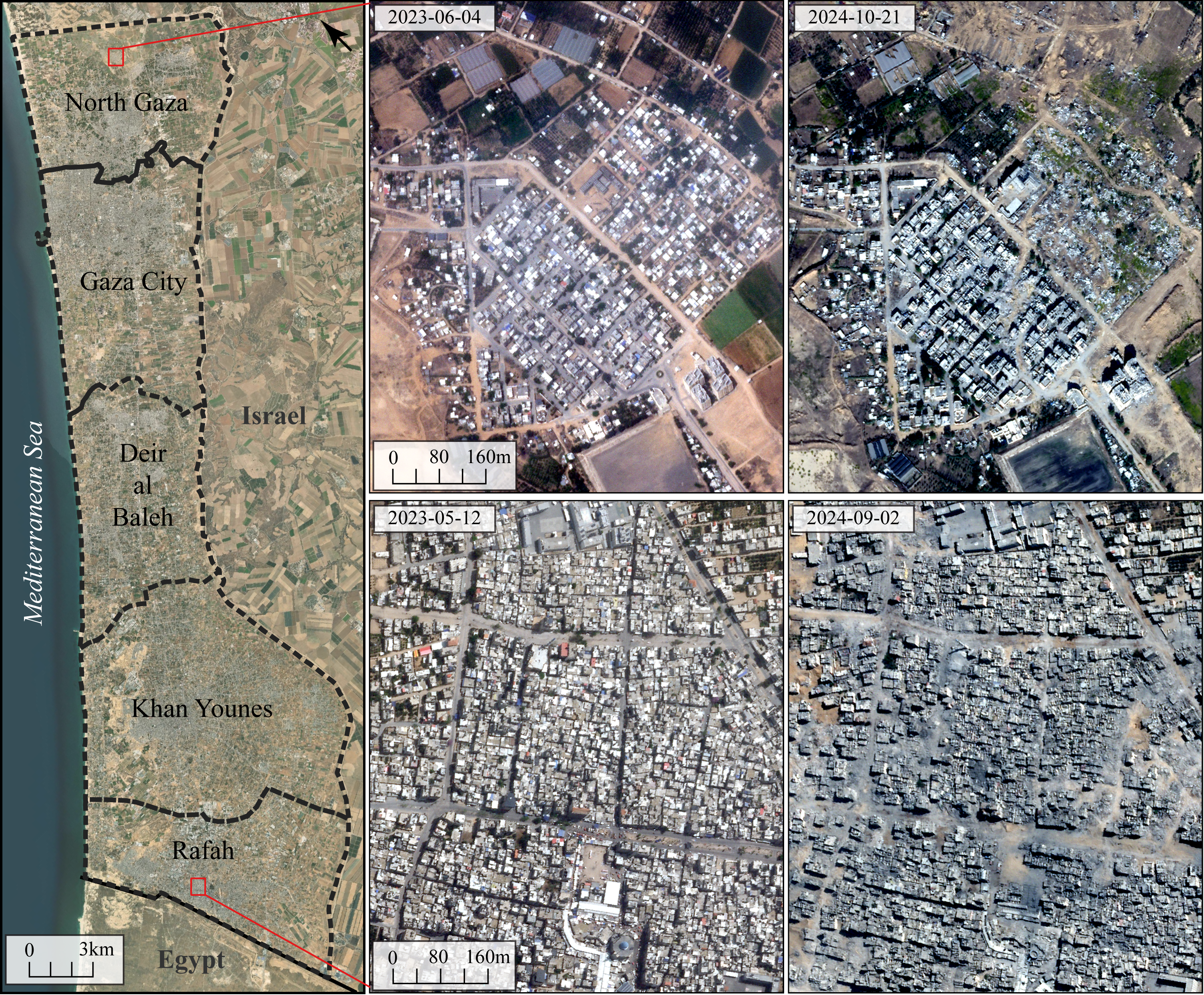}}
    \caption{Study setting. The left panel shows the study setting of the Gaza Strip with its five governorates. Inset panels (top) show VHR optical imagery before and during the war in the Beit Lahia area of Northern Gaza. Inset panels (bottom) show VHR optical imagery before and during the war in central Rafah.}
    \label{fig:setting_false_color}
\end{figure}


Armed conflict in Gaza has important parallels with other recent wars, which pose similar challenges in sustained urban damage monitoring. Like other conflict settings such as Ukraine and Syria, conflict in Gaza is characterized by widespread urban fighting -- and resulting damage -- to built-up areas. Difficulties accounting for damage across conflict settings involves a general lack of field-based data due to the dangerous (life threatening) situations on the ground. Conflict damage can occur over large spatial extents and long durations, which makes tasking of VHR imagery for detailed damage assessments difficult and introduces unique technical considerations associated with damage monitoring using medium resolution satellite data \cite{sticher2023toward, SCHER2025100217, VanDenHoek2021Sep}. The landscape of Gaza is unique to other conflict settings due to the densely built-up nature of its urban areas within a small spatial extent (365 km$^2$). The rapid and intense bombardment of the Gaza Strip over a realtively small area is also unique to other conflict settings, and underscores the need for active and sustained monitoring with low-latency to consistently capture the progression of damage over time. The high density and three dimensionality of Gaza's built-environment introduces geometric artifacts, distortions, and blind-spots in remote sensing imagery, challenging photo-interpretation of VHR optical imagery, which requires visibility around structures to classify forms of damage short of full or partial building collapse \cite{iwg_sem_guidelines}.


Urban damage due to armed conflict is increasingly assessed using remote sensing data. Academic literature on automated approaches for mapping armed conflict-induced building damage tends to be event-specific and rely on VHR sensor data, with case studies focused on individual towns or cities \cite{Bennett2022Sep}. The focus on commercial VHR data limits the degree to which Earth observation (EO) data can be widely and transparently used to better understand conflict impacts \cite{sticher2023toward, VanDenHoek2021Sep}. Interferometric synthetic aperture radar (InSAR) coherent change detection (CCD) approaches are well-suited for mapping damage in urban contexts due  to sensitivity in detecting structural shifts in built-up areas, where changes in complex radar signal characteristics can detect damage not always photo-interpretable in overhead optical imagery; regardless of cloud-cover or solar illumination conditions \cite{plank2014rapid}. While CCD has been applied extensively in disaster contexts \cite{plank2014rapid, ge2020review}, and in a handful of individual towns and cities affected by armed conflict (e.g., \cite{huang2023monitoring, boloorani2021post}), it has only once been applied to long-duration monitoring of conflict-induced damage across a nationwide extent with a case study in Ukraine \cite{SCHER2025100217}. This case study produced data on the timing and location of damage nationwide with three months latency to results. In Gaza, three months latency is not conducive to support decisions on the timescales of humanitarian recovery and response or reporting on the impacts of conflict by civil society and journalistic organizations. A sustained monitoring approach with low-latency is necessary to capture impacts from the fast-paced dynamics in this ongoing war.


In this study, we analyze 321 openly-accessible Sentinel-1 SAR images acquired before and during the war with a cloud-based InSAR processing workflow to produce over 3200 coherence images and monitor for indicated building damage using a long temporal-arc CCD (LT-CCD) approach. LT-CCD introduces formation of long temporal-arc InSAR pairing to CCD, drawing from knowledge on seasonal \cite{ferretti2002permanent} and geometric \cite{gabriel1989mapping} considerations for coherence estimation. LT-CCD constructs two stacks of single reference coherence images with matching temporal baselines across each stack and strict spatial (perpendicular) baseline criteria. Coherence image stacks for wartime and baseline periods are formed using single reference images acquired at similar times of the year and with minimal orbital offset (spatial/perpendicular baselines), secondary images acquired during pre-war periods, matching the distributions of temporal baselines for each stack, and reducing to stack statistics (mean and standard deviation) that are subsequently used for CCD \cite{stephenson2021deep, SCHER2025100217, Yang2024Jul}. We categorize damage by detecting acute and sustained reductions in coherence during the wartime monitoring period. We conduct damage monitoring during the wartime period and, to assess the occurrence of false positives, we conduct damage assessments "in reverse," using Sentinel-1 data for coherence estimation acquired two years prior to the conflict. We aggregate pixel-level damage data to 330,079 building footprints identified using pre-war VHR optical imagery \cite{hotosm_gaza} and report damage severity as a percentage of buildings damaged at each time step in monitoring. We compare LT-CCD damage with 928,397 damage event times and locations produced by the United Nations Satellite Center (UNOSAT) through visual interpretation of VHR satellite optical imagery across twelve dates between October 2023 and September 2024. We construct timelines of damage with LT-CCD and report spatiotemporal agreement with UNOSAT locations over time while examining reasons for agreement and disagreement. With LT-CCD, we map the destruction of Gaza over the first year of the conflict with approximately weekly temporal fidelity and report on the progression of damage over time as it relates to the progression of major conflict events. Preliminary results from this work were distributed to the global press and international humanitarian organizations over the first year of the war with low-latency and broad uptake, informing on the extent of damage in Gaza when commercial image providers limited data availability \cite{Tani2023Nov} and access to the field was restricted. We conduct this analysis with additional retrospective agreement exercises, more complete building footprint data, and full methodological transparency.

\section{Results}

\subsection{Agreement with UNOSAT}



\begin{figure}[H]
    \centering
    \centerline{\includegraphics[width=16cm]{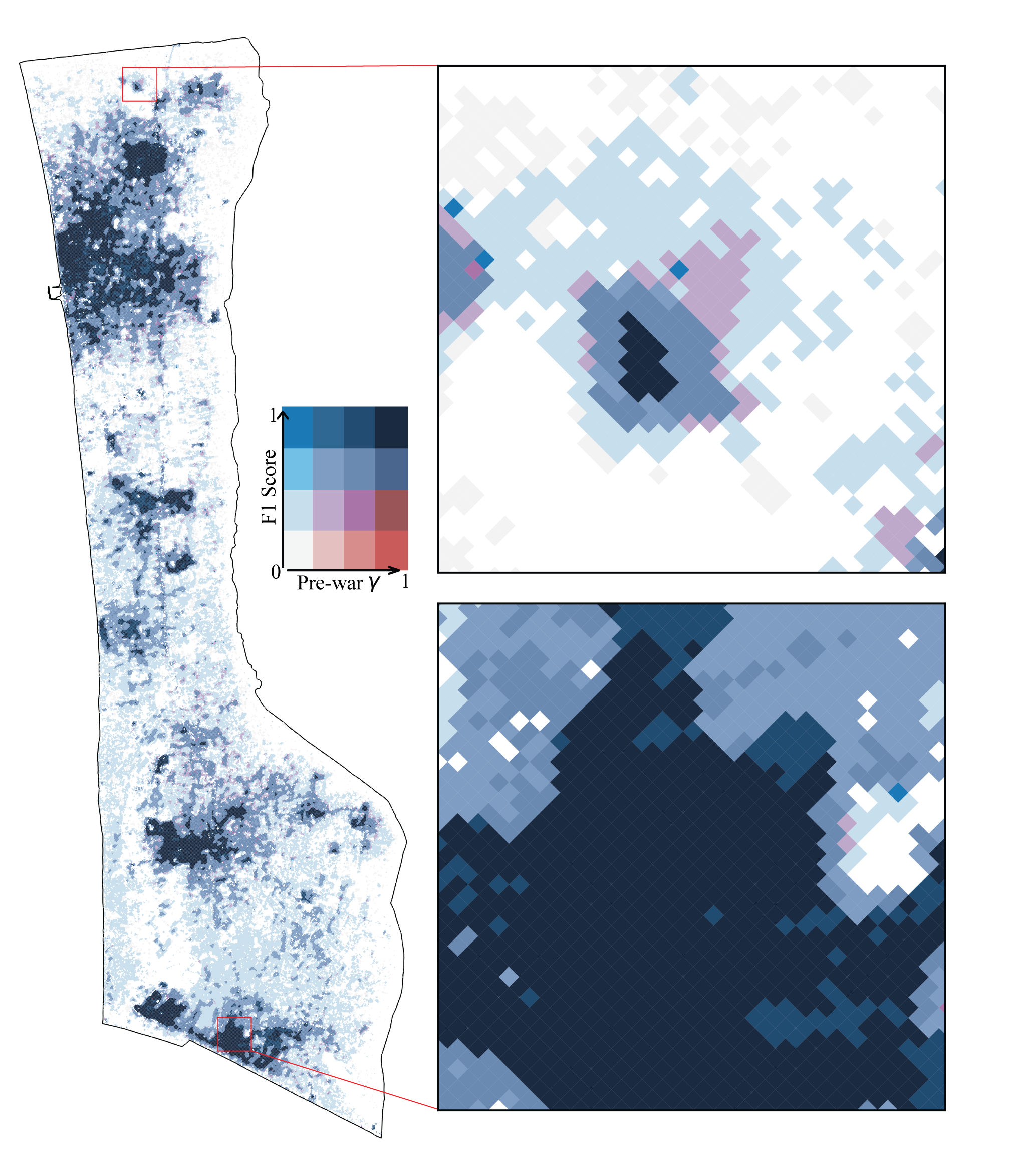}}
    \caption{Bivariate chloropleth of agreement (F1 score) between LT-CCD and UNOSAT by pre-war coherence ($\gamma$) magnitude. Areas styled in white are invalid for LT-CCD monitoring based on pre-war $\gamma$ characteristics.}
    \label{fig:unosat_agreement}
\end{figure}

\begin{figure}[H]
    \centering
    \centerline{\includegraphics[width=14cm]{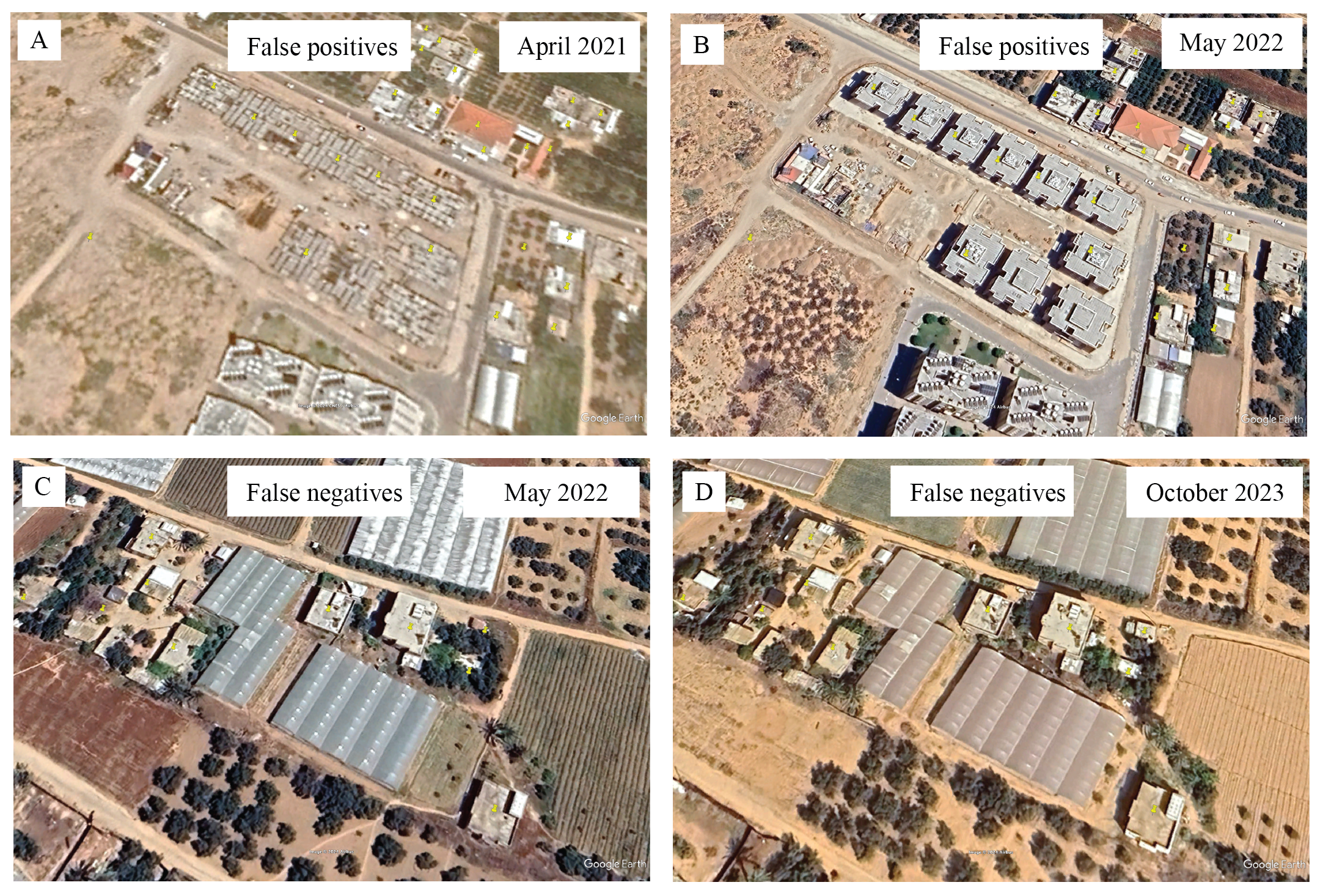}}
    \caption{(A) A region with false positives (31°24'17.93" N  34°22'07.77" E) in Google Earth basemap imagery acquired in April 2021. Note the open roofs and active construction of buildings. (B) The same region from Panel A imaged in May 2022. Note the completion of building rooftops. Between May 2022 and August 2023, solar panels were also added to many of these rooftops. These buildings register as false positives of damage during our counterfactual monitoring period because construction took place between counterfactual and pre-conflict baseline periods. (C) A region with false negatives (31°23'39.07" N  34°22'15.71" E) in Google Earth basemap imagery acquired in May 2022. Note buildings situated on managed agricultural lands with orchards, row crops, and greenhouses situated throughout the scene. (D) The same region from panel C imaged in October of 2023. Note changes visible in the landscape, such as the establishment of row crops, senescence of a vegetated crop canopy, and new greenhouse structural elements.}
    \label{fig:fn_fp_vignettes}
\end{figure}

We define areas valid for LT-CCD monitoring using pre-war coherence statistics (Methods Section 5.3.4) and detect 694,831 UNOSAT damage locations in areas valid for monitoring with an overall agreement of 92.5\%, true positive rate (TPR) of 86.2\%, false positive rate (FPR) of 1.2\%, F1 score of 91.8\%, and CSI of 85.2\%. For all UNOSAT times and locations of damage including those outside of areas valid for LT-CCD monitoring, our approach has overall agreement of 86.9\%, TPR of 74.8\%, FPR of 1.1\%, an F1 score of 84.8\%, and a CSI of 74.0\% (Supplementary Table \ref{tab:unosat_agreement_agg}). False positives are acceptably low and we capture the vast majority of UNOSAT locations over time. Agreement increases with the magnitude of pre-war coherence values (Fig. \ref{fig:unosat_agreement}). The median pre-war coherence value where true positives are detected is relatively high (0.69), whereas the median pre-war coherence value where we miss detection of UNOSAT locations is lower (0.45). UNOSAT locations that go undetected with our methods tend to occur in areas where SAR signal variability drives lower pre-war baseline coherence magnitudes, which is related to the pre-war built-up density, construction activity, and land management practices (Fig. \ref{fig:fn_fp_vignettes}). The median built-up area percent where we detect UNOSAT locations at the pixel level is 34\%, whereas the median built-up area where we miss UNOSAT locations is 19\%. In total, UNOSAT reported 163,778 locations of damage as of 6 September 2024, the last UNOSAT survey during the study period. By evening local time on 5 September 2024, we detect 175,691 locations of building damage in Gaza. Because UNOSAT does not release its reference building footprint data and only produces point locations to represent damaged buildings, we cannot determine whether this difference in aggregate damage is due to how UNOSAT defines building outlines or whether this is due to other factors.

\subsection{Damage over time}

\begin{figure}[H]
    \centering
    \includegraphics[width=14cm]{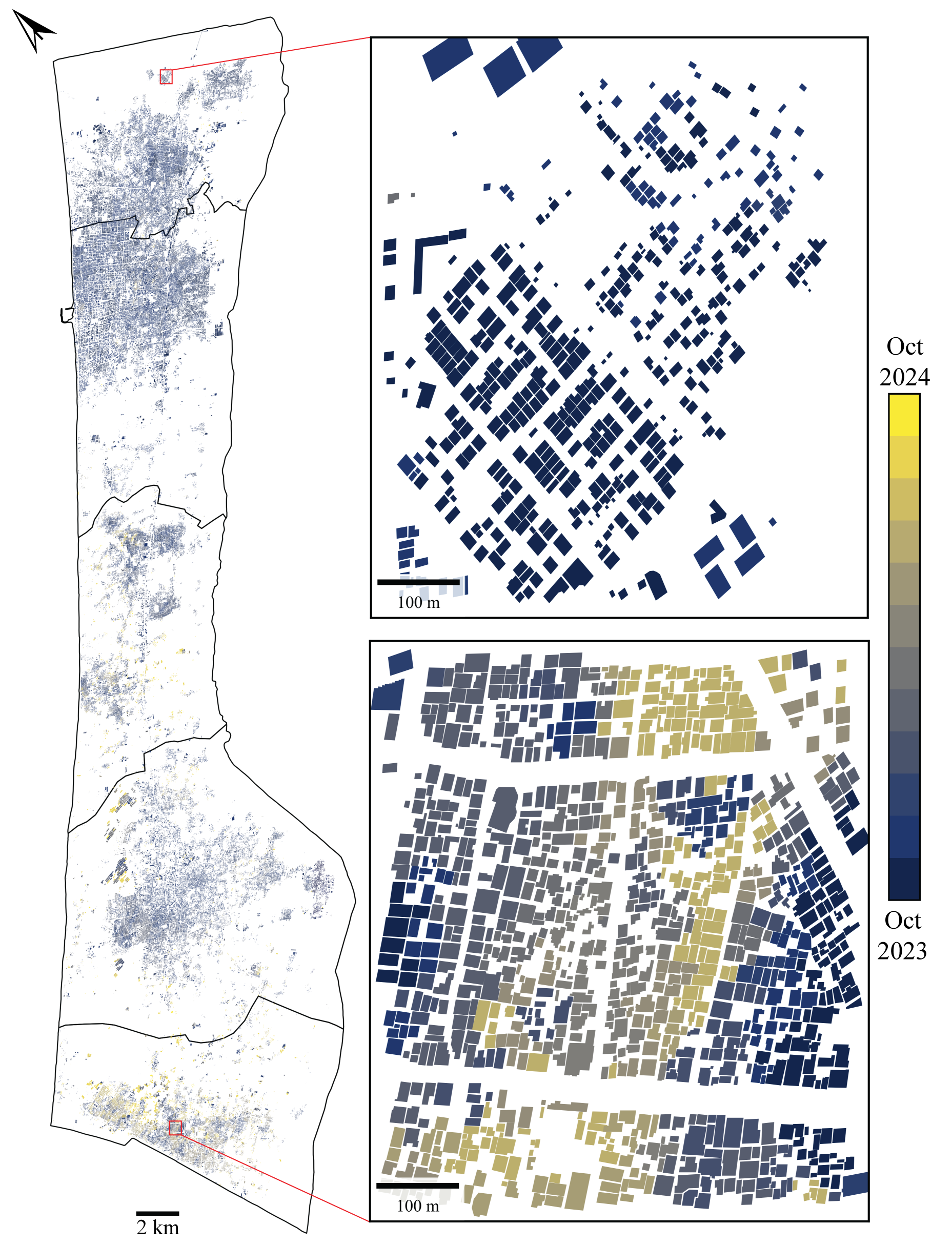}
    \caption{Aggregate damage detected across the duration of the study period with building footprints colored by the month where damage is first detected.}
    \label{fig:results}
\end{figure}
\begin{figure}[H]
    \centering
    \includegraphics[width=16cm]{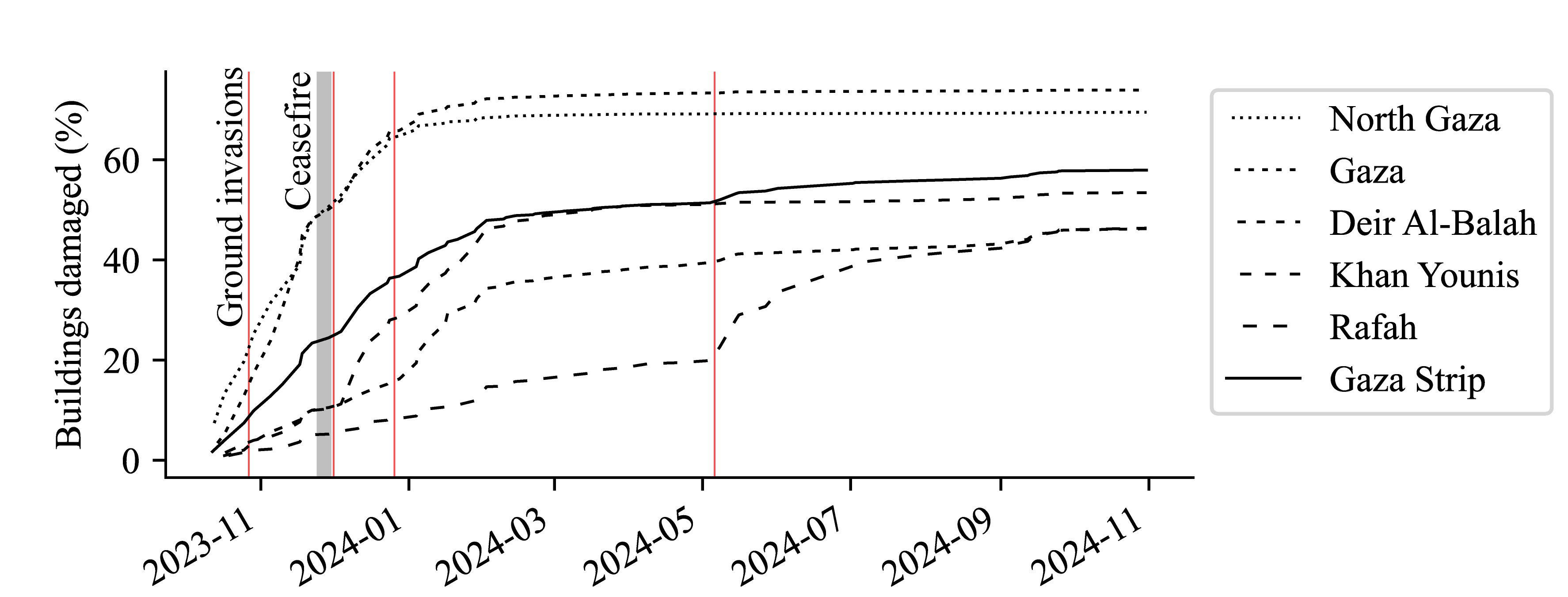}
    \caption{The percentage of buildings damaged over time. Cumulative damage is aggregated across each governorate and the Gaza Strip as a whole. Ground invasions are indicated with vertical red lines. The invasions of North Gaza and Gaza City begin on 27 October 2023. The 24-30 November 2023 temporary ceasefire is shaded in grey. The beginning of the battle of Khan Younes is marked in red on 1 December 2023 along with the encircling of Deir al-Baleh on 26 December 2023, and the invasion of Rafah on 6 May 2024.}
    \label{fig:results_timeline}
\end{figure}

Damage detection at weekly temporal fidelity reveals regional conflict dynamics (Figs. \ref{fig:results} and \ref{fig:results_timeline}). For the first 6 weeks of the war and before the temporary ceasefire, we detect on average 2,300 new buildings damaged per day, with damage mainly focused in the northern areas of Gaza. At the war's onset, aerial bombardment was focused in North Gaza and Gaza City, followed by Israeli military ground invasions beginning on 27 October 2023. In the North Gaza and Gaza City governorates, over half of all buildings were damaged by the end of November 2023. During this phase of the war, ground invasions were limited to these two northern governorates while airstrikes took place across all governorates including in the central and southern areas of Gaza.

In the last week of November was a six-day temporary ceasefire and hostage exchange from 24-30 November 2023 \cite{ceasefire_start_isw, ceasefire_end_isw}. Relative calm during the temporary ceasefire is evidenced by a 78\% decrease in the rate of new damage detected compared to the first six weeks of the war across all of Gaza (Table \ref{tab:decrease_during_ceasefire}). In the southernmost Khan Younes and Rafah governorates, the decrease in new damage detected during the temporary ceasefire is slightly more pronounced than in other governorates. Because Sentinel-1 acquired data on 22 and 29 November 2023, we were able to capture this notable slowdown in new damage during the brief respite from fighting. Due to the timing of Sentinel-1 acquisitions, we do not detect a total pause in damage, as airstrikes and fighting continued from the 22nd to the 24th of November \cite{raleigh2010introducing} and Sentinel-1 did not make acquisition to precisely capture the six-day ceasefire period.

By the end of the temporary ceasefire, 10.5\% of Deir Al-Baleh, 10.3\% of Khan Younes, and 5.2\% of Rafah were damaged. Rates of new damage increase markedly in Khan Younes corresponding with the onset of intensified bombardment and an Israeli military ground offensive that captured the city center in the first week of December \cite{Barbakh2023Dec}. Shortly after the Israeli capture of Deir Al-Baleh, the Israeli military announced that it was pressing into the adjacent city of Khan Younes \cite{Ravid2023Dec}, where we detect a notable increase in new damage beginning in January 2024 when the Israeli military fully encircled the city after weeks of battle \cite{Bigg2024Jan}. By mid-March 2024, over half (50.1\%) of buildings in Khan Younes and 37.5\% of buildings in Deir Al-Balah were damaged. At this time, the southernmost Rafah governorate, which had been a target of aerial bombardment throughout the conflict, had 17.8\% of buildings with damage but was the only remaining governorate where an Israeli ground invasion had not occurred. New damage detection in Rafah increases markedly following the beginning of an Israeli military ground invasion on May 6, 2024. Less than two months later, the percentage of buildings damaged in Rafah more than doubled to 40.9\%.

\begin{table}[H]
\begin{center}
\caption{Percent decrease in the rate of new damage detected during the temporary ceasefire relative to the first six weeks of the war.}
\label{tab:decrease_during_ceasefire}
\begin{tabular}{lc}
\toprule
 Region & Decrease (\%) \\
\midrule
North Gaza & 72 \\
Gaza & 80 \\
Deir Al-Balah & 76 \\
Khan Younis & 86 \\
Rafah & 91 \\
\midrule
\textbf{Gaza Strip} & \textbf{78} \\
\bottomrule
\end{tabular}
\end{center}
\end{table}

In aggregate, LT-CCD classifies 57.9\% (191,263) of all pre-war OSM buildings mapped in Gaza as likely damaged or destroyed through the first year of the war (Fig. \ref{fig:results}). 69.5\% of buildings in North Gaza, 73.9\% in Gaza City, 46.2\% in Deir al Baleh, 53.5\% in Khan Younes, and 46.4\% in Rafah were likely damaged. Areas with low magnitude, variable pre-war coherence, corresponding to areas with low built-up density, were classified as invalid for monitoring, constituting 14.4\% (47,516) of OSM building footprints (Table \ref{tab:lc_valid_invalid}). As a percentage of building footprints in areas valid for LT-CCD monitoring, over two thirds (67.8\%) of all buildings in Gaza were damaged.

\begin{table}[H]
\begin{center}
\caption{Mean pre-war $\gamma$ and OSM built-up density in areas valid and invalid for LT-CCD monitoring.}
\begin{tabular}{lcc}
\hline
 LT-CCD & Pre-war $\gamma$ & OSM built area (\%) \\
 \hline
 \hline
 Valid & 0.50 & 18.9 \\
 Invalid & 0.24 & 2.7 \\

\bottomrule
\end{tabular}
\label{tab:lc_valid_invalid}
\end{center}
\end{table}

\subsection{Accounting for confounding factors}

\begin{figure}[H]
    \centering
    \centerline{\includegraphics[width=1.2\linewidth]{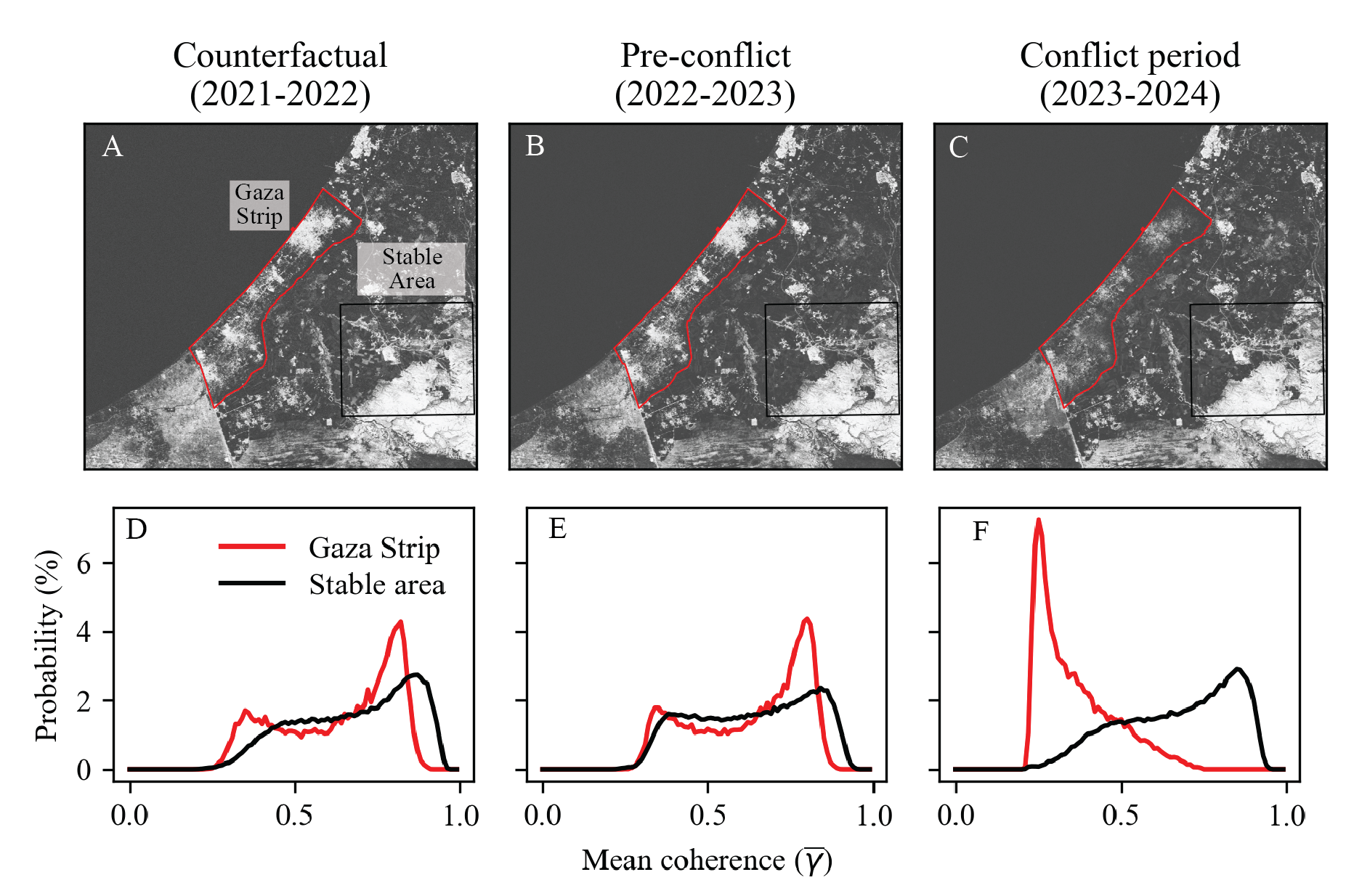}}
    \caption{Average coherence across all three monitoring epochs. (A) The average coherence magnitude ($\overline{\gamma}_{counterfactual}$) across all coherence grids used in the counterfactual monitoring period. The Gaza Strip is outlined in red and the pseudo-stable area in Israel is outlined in black. (B) Average coherence across all images used during the baseline pre-conflict period ($\overline{\gamma}_{pre}$). (C) Average coherence across all images used for monitoring during the conflict period ($\overline{\gamma}_{conflict}$). (D, E and F) Probability distributions of coherence values for the Gaza Strip and the pseudo-stable area in Israel pertaining to data in panels A, B, and C respectively. }
    \label{fig:coherence_distributions}
\end{figure}

We assess the overlap in probability distributions between all three epochs in monitoring (pre-war baseline, counterfactual, and wartime periods) using the Hellinger distance metric. There is substantial, approximately 90\%, overlap within Gaza and in southern Israel in coherence value distributions assembled outside of wartime periods (Fig. \ref{fig:coherence_distributions}, Table \ref{tab:hellingers}). In southern Israel, distributions of coherence values have substantial overlap across all three epochs where coherence estimates were formed, indicating a lack of broad changes in coherence outside of Gaza during the war. In contrast, coherence distributions within the Gaza Strip show a marked difference during the wartime monitoring period relative to the pre-war and counterfactual periods with Hellinger distances over 0.65, indicating that two thirds of the probability distributions no longer overlap with pre-war coherence data. This is indicative of the extensive damage to built-up areas characterized in this study.

\begin{table}[H]
\begin{center}
\caption{Hellinger distances between mean coherence during the pre-war baseline, conflict monitoring, and counterfactual periods within Gaza and in the southern desert of Israel.}
\label{tab:hellingers}
\begin{tabular}{lrr}
\toprule
 Monitoring epochs & Gaza & Southern Israel \\
\midrule
Pre-war and counterfactual & 0.08 & 0.13 \\
War and counterfactual & 0.64 & 0.08 \\
War and pre-war & 0.66 & 0.09 \\
\bottomrule
\end{tabular}
\end{center}
\end{table}


We test for sustained decorrelation after initial damage detection by accounting for the time between initial and secondary confirmation of damage signals. We find that 79.5\% of initial damage detections are confirmed within one month (31 days) of the first date of detection. Initial damage detection is most frequently confirmed (63.6\% of the time) within 12-days, which is the temporal lag time over which Sentinel-1 acquires imagery with the same orientation of illumination. On any given monitoring date, damage detected that remains unconfirmed within one month does not exceed 3 km$^2$ in area, amounting to 3.2\% of damage detected at that date in monitoring. While time to confirmation of initial damage varies, 99\% of all initial damage is confirmed within 176 days and all initial damage is eventually confirmed by the end of the study. We report agreement metrics with UNOSAT above before applying any persistence criteria, and the false positive rate is acceptably low as a result.

\begin{figure}[H]
    \centering
    \includegraphics[width=10cm]{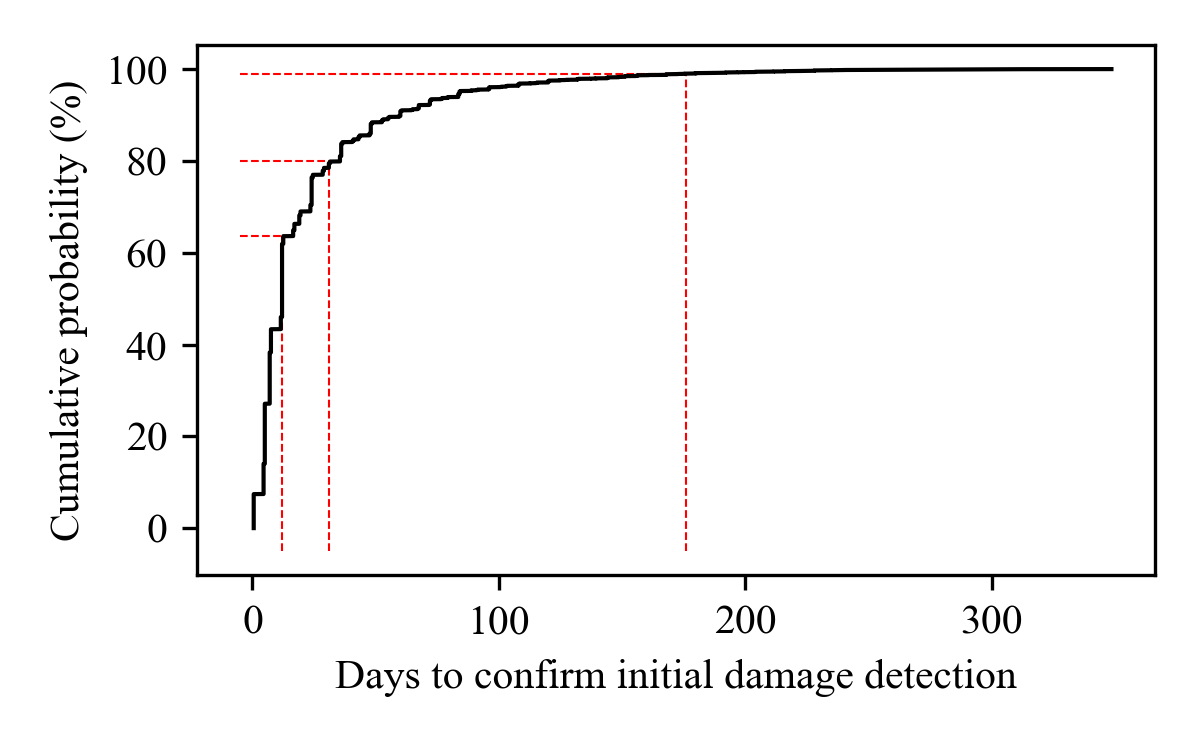}
    \caption{Cumulative distribution function of the number of days between initial and secondary damage detection. The horizontal and vertical dashed red lines indicate the probabilities that initial damage is confirmed after 12 (63.6\%), 31 (79.5\%), and 176 days ($>$99\%).}
    \label{fig:persistence_of_damage}
\end{figure}

\section{Discussion}

Most of the damage detected in this study occurs within the first nine months of the war. The first three months of the war had the highest rates of new damage. This pace of damage detection has almost no precedent in peer-reviewed literature on EO-based monitoring of damage during aerial bombardment campaigns aside from the most severely-affected towns and cities in Ukraine damaged during the 2022 Russia-Ukraine conflict \cite{SCHER2025100217}. To our knowledge, the severity of damage in Gaza compares only the some of the most severe wartime damage in modern history dating back to WWII. In the German city of Dresden, for example, 54\% of buildings were damaged or destroyed due to Allied bombardment \cite{tustin130523dresden}. Across 51 German cities subject to Allied bombardment from 1942-1945, 40-50\% of urban areas were damaged or destroyed, amounting to an estimated 10\% of the total German building stock \cite{Pape1996}. In Gaza, the entire enclave has a damage severity comparable only to some of the most heavily damaged localities in the Russia-Ukraine conflict or WWII. This severity of damage, however, occurs across the entire extent of the Gaza Strip, which means that this damage is not just severe in some localities but it is inflicted across the broader administrative unit.

Geographic scales of damage relate to the pixel resolution EO data for wartime damage monitoring. The pixel resolution of data used for LT-CCD here is consistent with the radii of building destruction from one of the most commonly-deployed bombs in Gaza \cite{Bott2023Dec}: the Mark 82 munition, which carries 89 kg of explosive \cite{gichd_mk82}. Detonation of the Mark 82 bomb on a flat surface will collapse most buildings and severely damage concrete structures within a 31m radius; an area approximately two times the 40m pixel spacing used in this study \cite{gichd_mk82}. The total number of bombs, by type, dropped in Gaza is not publicly known. By mid-December 2023, a United States intelligence leak reported in the media stated that 29,000 air-to-ground munitions had been deployed in Gaza \cite{Bertrand2023Dec}. Hundreds of these air-to-ground munitions were much more destructive and lethal one ton (Mark 84) bombs \cite{Bertrand2023Dec, Bott2023Dec, Qiblawi2023Dec}. As a back-of-the-envelope estimate, an area of 88 km$^2$ would be damaged or destroyed if all munitions were Mark-82 and did not overlap. By mid-December 2023, we detect 79 km$^2$ of damaged area gridded at 40m pixel spacing. An approach to map damage at 30-50 centimeter scale, akin to UNOSAT-style assessments, is not as relevant to the geographic scale of conflict damage processes, which can be captured persistently over time with this medium-resolution approach. Within areas with sufficiently high pre-war coherence characteristics, our approach drastically reduces latency from weeks for UNOSAT assessments to hours with LT-CCD for more timely and actionable data that can subsequently prioritize tasking of follow-on VHR imagery for higher-fidelity assessment of damage at the level of individual buildings.

The net difference in the number of buildings detected by our methods versus UNOSAT could be explained by the difference in pre-war reference building footprint vector data, of which UNOSAT does not make publicly available, or it could be due to detection of damage in densely built-up areas. UNOSAT-style methods are known to lack sensitivity when damage occurs without evidence readily visible at the rooftop in VHR imagery, with errors of omission of 26.8\% where UNOSAT data have been validated in one post-earthquake setting \cite{Rathje2011Oct}. Examples of lateral damage, including plastic deformation of buildings, residual drift, progressive collapse \cite{Ngo2007Jan}, and damage to structural facades \cite{O'Donnell2024} are instances where objects that scatter radar echoes may lead to CCD detection of damage that UNOSAT omits \cite{plank2014rapid}. Since neither LT-CCD or UNOSAT data are field validated, we can consider that LT-CCD may be capturing data that UNOSAT omits in areas where visibility is occluded, while LT-CCD with C-band SAR is expected to miss damage in areas where mixed-pixel landcover effects on radar scattering drives lower coherence.

Merging LT-CCD with VHR optical data can form more complete EO-based damage estimates by considering the strengths of CCD for capturing lateral forms damage \cite{O'Donnell2024} in areas with high pre-war coherence and UNOSAT capturing damage in less densely-built areas with more variable radar scattering. We merged unique building locations identified by LT-CCD and UNOSAT to develop a more comprehensive assessment of built-up area damage. In total, over 214,000 buildings are likely damaged or destroyed at the end of the study period, constituting 63.99\% of all OSM mapped buildings. This represents an upward revision of 12\% from our initial estimates and a marked 31\% increase from the UNOSAT estimates. Public release of reference building footprint data used by UNOSAT would enable transparency and reproducibility of damage estimates for assessing strengths and limitations of either approach, but the present lack of publicly-available reference building footprint data from UNOSAT limits independent assessments of differences in each approach at the individual building footprint level.

It is difficult to untangle the multivariate drivers on long temporal-baseline coherence and factors affecting the detectability of damage. Vegetation lowers coherence generally \cite{rs12162545} and the resulting ability to detect damage. Buildings with larger footprints are more reliable to monitor for damage than smaller buildings \cite{natsuaki2018sensitivity}. Variation in the orientation of an urban grid can change dominant scattering mechanism from double-bounce to volumetric scattering \cite{DelgadoBlasco2020Jan}. Volume scattering in urban areas may help to explain the sensitivity of CCD approaches to lateral damage to building facades \cite{O'Donnell2024}. Disambiguating the different influence of coherence signal drivers such as vegetation presence, built-up density, sensor, urban grid orientation, scattering mechanisms, and seasonal scattering variability can contribute toward better understanding of drivers of urban coherence characteristics over space and time. Characterizing these multi-variate drivers of damage detectability can help to better understand the detectability of coherence across geographic settings.

The occurrence of repeat damage - where an area is struck more than once over time - may register as damage initially but, as a war continues, more strikes may occur in the same image region ostensibly increasing the severity of damage within that pixel. Reports from the news media document increasing damage over time, with one example from the British Broadcasting Service (BBC) \cite{ByDominicBailey2023Dec}. Using photos from the ground of the same area over the first six weeks of the war, the BBC highlights a mosque and surrounding buildings with increasing levels of damage visually evident in Beit Lahia, North Gaza. Our methods detect initial damage at this site but, as the site is struck more over time, increasing severity of damage is not detectable with our methods due to the binary nature of damage classification. Conceptualizing a notion of increasing damage severity at the level of individual pixels will be important to capture signals of repeat damage in future and ongoing work; and may be a path forward given the limitations of medium-resolution SAR for capturing individual building-level damage severity \cite{natsuaki2018sensitivity}.

While opportunities remain for increasing the fidelity of CCD approaches for mapping repeat and increasing severity of damage across geographic settings, the LT-CCD approach operationalized in this study offers several advantages over bi-temporal and fixed-temporal baseline CCD approaches developed for disaster contexts (e.g., \cite{stephenson2021deep}). With LT-CCD, all coherence estimation is conducted using secondary SAR imagery acquired before the war onset, which makes testing for persistent decorrelation possible \cite{SCHER2025100217}. A fixed-temporal baseline approach agnostic to the conflict onset precludes the ability to test for signals of persistent decorrelation relative to a pre-war period. LT-CCD also allows for strict considerations of spatial baselines across InSAR pairs used for coherence estimation. With LT-CCD, we can restrict InSAR pairing to acquisitions with short spatial (perpendicular) baselines and help to mitigate potential effects of spatial decorrelation due to large differences in the point of illumination along the SAR orbit track \cite{gabriel1989mapping}. Additionally, bi-temporal CCD approaches commonly apply an approach for "histogram matching" of coherence values for pre- and post-event coherence estimates \cite{Yang2024Jul, O'Donnell2024}. With such widespread damage in Gaza and a relatively small area of interest, a presumption of stability to alter coherence values during the conflict isn't necessarily appropriate. Instead, we opt to match the histograms of temporal baselines used for InSAR image pairing at each timestep in monitoring and for each stack of coherence images reduced to central tendencies and used for CCD (Methods Fig. \ref{fig:baselines}). We find that the formation of image stacks using reference images acquired around the same time of each year, at a similar orbital vantage points, and secondary images with the same distribution of temporal baselines does well to create similar coherence estimates when reduced to central tendencies, as expressed through the Hellinger distance comparisons above.

Another important benefit of the LT-CCD approach is to mitigate for signals of wartime construction. In Gaza, as the Israeli military conducted ground invasions, it also built military checkpoints and fortifications, including in areas it termed the Netzarim and Philedelphi corridors in central and southern Gaza. Because our long temporal-arc approach for estimating coherence only utilizes secondary images from the pre-war period, we are only sensitive to tracking areas with long-term pre-war coherence characteristics conducive to CCD. This means that, if an area is damaged during the war and subsequently built-up with military fortifications, these military installations should not falsely register as damage because decorrelation of SAR signals relative to the beginning of the war has already been mapped as damage.

As the war drags on, the gradual decay of coherence over time will lower the magnitude of mean coherence across an image stack and across the region (e.g., \cite{kellndorfer2022global}). Because the Israel-Hamas war is now in its second year, and because we have demonstrated that LT-CCD is robust to temporal decay phenomena for damage detection over a year of monitoring, this approach can be operationalized for monitoring longer than one-year durations by adjusting the pre-war benchmarking period after each year of monitoring. For example, pre-war reference periods for coherence estimation after the first year of fighting can shift to the second year, and all new damage can be reported as it was detected in the second year of monitoring as a strategy to mitigate for increasing temporal baselines. 

The importance of continued monitoring, regardless of the reported state of the conflict, is not only integral to capture damage but also to monitor ceasefire effectiveness. In November of 2023, we capture a slow-down in new damage detection during the six-day temporary ceasefire and hostage exchange. This underscores another important factor for high temporal fidelity and active monitoring of damage during ongoing armed conflicts. As ceasefire agreements are put in place, approaches like ours can monitor for signals representative of new damage during ceasefire periods. Physically-based and transparent data produced agnostic of any political actors on the ground is important to document potential violations of ceasefire agreements.

These damage estimates can be generated within hours of a Sentinel-1 image down-link. The new temporal fidelity of rapidly-generated data on wartime damage to built-up areas is necessary for timely decision support at humanitarian organizations coordinating civilian recovery and response activities in war-torn areas. Timely insights enable public understanding of conflict impacts as communicated through journalistic organizations, to assess knock-on effects such as rubble accumulation, dust exposure, conflict-induced migration, and to characterize broader long-term environmental implications of armed conflict. Low-cost approaches utilizing open source EO data can improve transparency and reproducibility and contribute to the better use of EO data for more holisitically understanding the landscape impacts of war and conflict.

\section{Methods}
\subsection{Geography}

\begin{figure}[H]
    \centering
    \centerline{\includegraphics[width=0.8\textwidth,keepaspectratio]{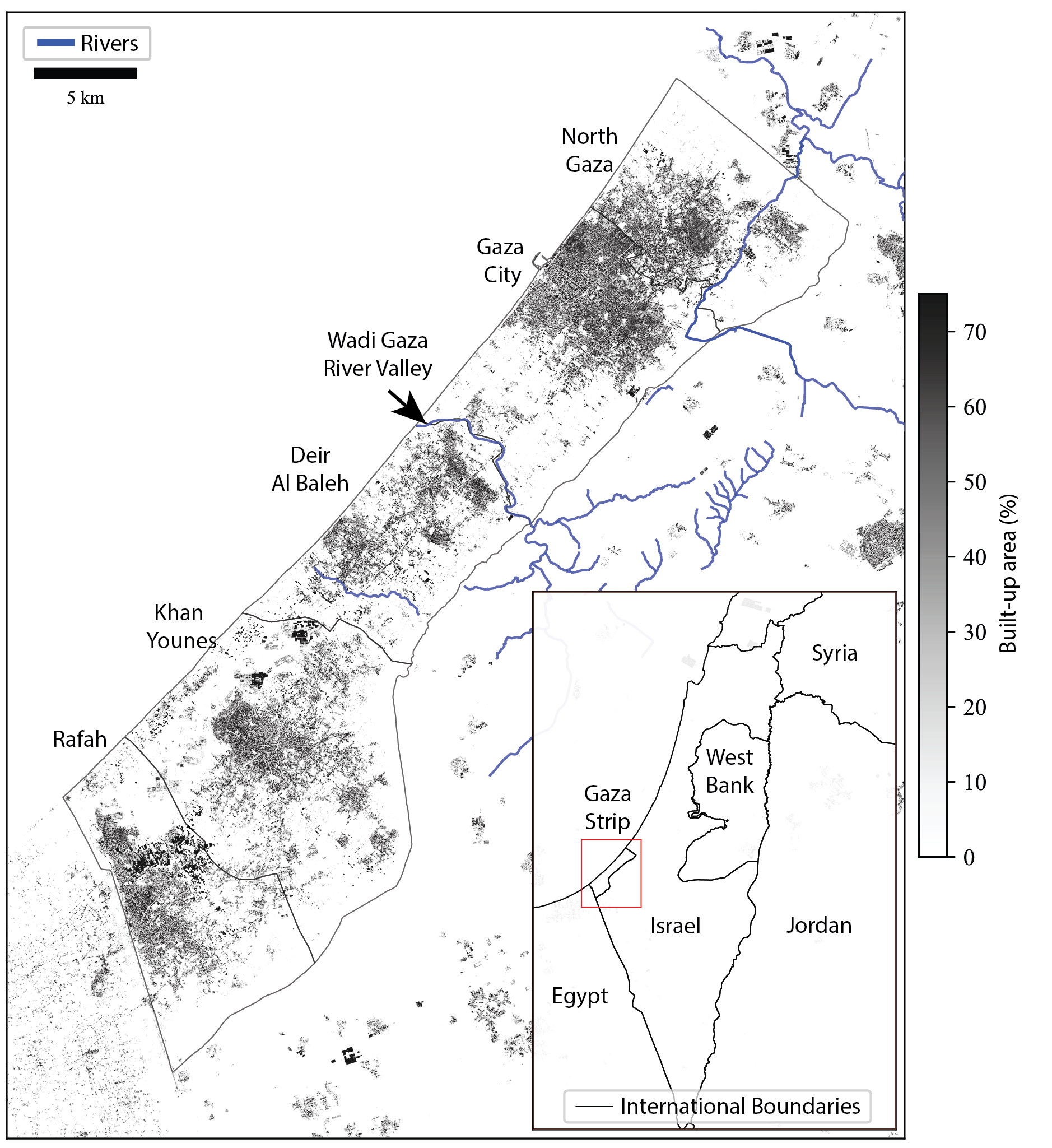}}
    \caption[The Gaza Strip and its five governorates]{The Gaza Strip and its five governorates outlined over data on built-up density from the 2023 Global Human Settlement collection \cite{Pesaresi2023}. An inset map in the lower right panel outlines the inset zoom in red and includes recognized international boundaries \cite{Runfola2020Apr}.}
    \label{fig:setting}
\end{figure}

The Gaza Strip (Gaza) is located on the eastern boundary of the Mediterranean Sea and is besieged \cite{gaza_blockade} by Israel to the north, east and west and bordered by Egypt to the south (Fig. \ref{fig:setting}). There are approximately 2.2 million people in Gaza as of 2023 distributed across an area of 365 km$^2$, making the enclave one of the most densely-populated areas in a continental setting on Earth \cite{wb_pop_density}. In 2015, 14.56\% of the Gaza Strip (43.59 km$^2$) was built-up with human settlements \cite{Pesaresi2023}. The average density of built-up area per hectare is 19.8\% with a standard deviation of 1,800 m$^2$ \cite{Pesaresi2023}. Intermixed perennial and annual vegetation is distributed within otherwise densely built-up areas of the Gaza Strip. On the periphery of densely built-up urban areas are agricultural regions. In peri-urban and rural areas outside of major urban centers, agriculture and farming are the primary land uses \cite{UNCTAD2015}. These agricultural regions include infrastructure such as greenhouses within and between agricultural plots.

\noindent

\subsection{Climate and Environment}

Gaza is in a Mediterranean climatic environment consisting of generally warm and dry summer months with average daily surface air temperatures above 25ºC from June - August \cite{worldbank2023}. Rainfall occurs from October through May, with the most average annual rainfall arriving between December and January \cite{worldbank2023}. Rainfall each year accumulates to 35 cm on average \cite{worldbank2023}. Following seasonal precipitation events, there is green-up of annual vegetation that contrasts with the dry season months where greenness is limited to otherwise sparse evergreen perennial vegetation. Governorates to the south are slightly more arid than the north. Episodic, annual rainfall occurs in events that cause surface water runoff, flooding, and erosion \cite{Al-Najjar2022Oct}. Together, the annual green-up of vegetation \cite{santoro2009signatures}, changes in soil moisture, episodic rainfall \cite{chaabani2018flood}, and erosion \cite{schepanski2012evidence} will drive changes in radar scattering characteristics that may confound the detection of damage.

\subsection{Characteristics of urban damage}

War-related damage to buildings in the Gaza Strip has several causes including due to direct aerial impact, ground-based fighting, controlled-demolitions \cite{Abraham2024Jul}, and damage from blast waves and ejecta from nearby detonations \cite{Sorensen2011Apr}. Aerial bombardment has reportedly included the use of thousands of conventional Mark-82 227 kg bombs \cite{Bott2023Dec}, which, when exploded at the surface, have a blastwave radius where unimpeded structures within 31 m are destroyed due to combined effects from overpressure as blastwaves propagate and the resulting under-pressure in the blastwave wake \cite{mk-82-blast, Shirbhate2021May}. Heavier 500 kg Mark-83 and 1,000 kg Mark-84 bombs have also been dropped across Gaza at least several hundred times \cite{Qiblawi2023Dec, Stein2023Dec}, delivering much more destructive force with wider blast and lethality radii than the Mark-82 \cite{Bott2023Dec}.

\subsection{Data}
\subsubsection{Sentinel-1 interferometric coherence data products}

InSAR coherence ($\gamma$) \cite{zebker1992decorrelation}, a derivative of common InSAR processing, is often used to assess the reliability of InSAR measurements for phase change analysis applied to elevation or deformation mapping \cite{hao2008analyzing, hooper2008multi, hooper2012recent, Shanker2011Jan}. Coherence has also become a metric used for operational building damage mapping in disaster contexts, but can be prone to non-damage drivers of coherence loss (decorrelation). When sensor and dielectric changes do not limit the reliability for monitoring areas with otherwise stable radar scattering characteristics, $\gamma$ is sensitive to subtle types of building damage not always visible in overhead optical imagery \cite{cho2023backscattering}. As a result, $\gamma$ has become the metric utilized by civilian spaces agencies and research groups to map damage from geohazard and extreme weather events with bi-temporal (before/after) change detection \cite{Ge2020Apr}. 

We use data from the Sentinel-1 constellation \cite{sentinel1} to map indicators of damage. The constellation, with the first launch of Sentinel-1A in 2014, is a C-band (5.405 GHz, 5.6 cm wavelength) sensor and constitutes the first regular repeat satellite SAR system with openly-accessible data freely available to the public. Sentinel-1B was launched in 2016 and, together with Sentinel-1A, generated a record of 6-day repeat acquisitions across most land areas on Earth. Sentinel-1B was decommissioned in December of 2021 following failure of onboard equipment and left Sentinel-1A as the only satellite in the constellation to acquire data during the Israel-Hamas war. Nonetheless, Sentinel-1B acquisitions prior to the onset of the 2023- Israel-Hamas war are still important sources of data for coherence estimation.

Beginning in April of 2024, a thruster anomaly on Sentinel-1A caused mission controllers to relax the precision of an orbital tube diameter leading to larger than normal ground track deviations in acquisitions \cite{billhauer22024Apr}. This drove an increase in perpendicular baselines for some acquisitions upwards of several hundred meters beyond average \cite{Sentinel2024}. With perpendicular baselines greater than several hundred meters, coherence changes over urban areas can be dominated by the influence of spatial baseline decorrelation \cite{Grey2003Sep}. While Sentinel-1 acquired 97 images in total during the conflict period, we limit our monitoring using the 61 Sentinel-1 images described above and omit the rest of the record due to a combination of large ground-track deviations from average and finite computing resources. Nonetheless, the average frequency of temporal revisit over Gaza using these 61 images is conducive for pseudo-weekly monitoring of indicated damage across the first year of conflict.

The Alaska Satellite Facility's Hybrid Pluggable Processing Pipeline (HyP3) \cite{hyp3} is an on-demand cloud-based SAR processing API that includes interferometric processing with an instance of the GAMMA software \cite{gamma}. We tasked HyP3 with custom job lists to process interferometric data products and generate gridded coherence estimates used in this study. Cloud-based InSAR processing tools like HyP3 enable scaling of CCD approaches with multitemporal stacks of data available at each time step in monitoring, which would otherwise not be feasible to conduct in a timely manner with the need to download and process dozens of images on a typical desktop computer. Interferometric processing at HyP3 allows users to choose the window size within which coherence is estimated across neighboring pixels, where 10x2 pixels in range and in azimuth (radar geometry) is the highest resolution available. This results in coherence data products with 40m pixel spacing and 80m nominal resolution, which we employ in this study. More information on the Sentinel-1 InSAR coherence products that we utilized in this study is available in the Hyp3 documentation \cite{hyp3}.

\subsubsection{HOTOSM Building Footprint Data}
\begin{figure}[H]
    \centering
    \includegraphics[width=12cm]{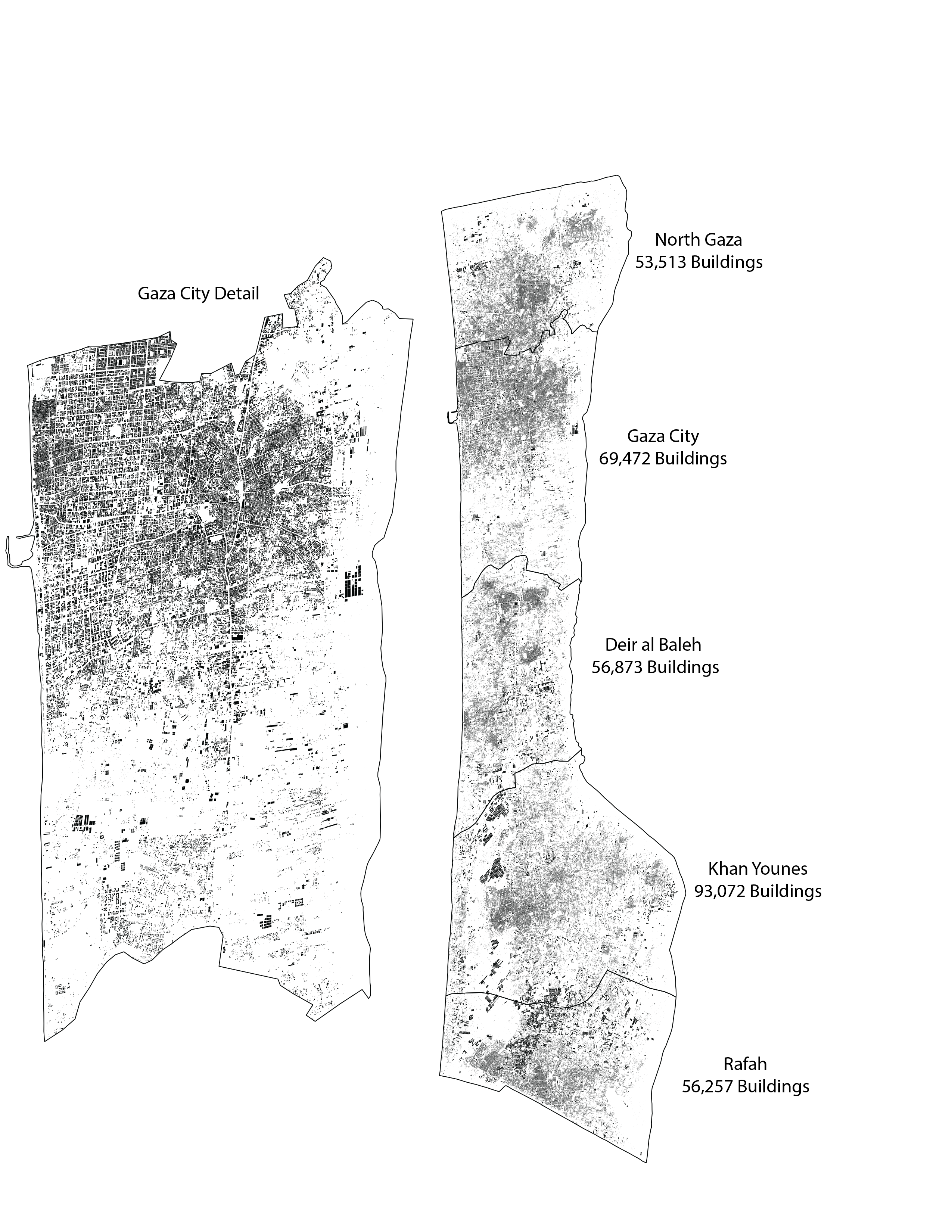}
    \caption{OSM building footprints with the count of building footprints in each governorate.}
    \label{fig:footprints}
\end{figure}

Building footprints are used to restrict monitoring to regions with built-up structures known to be present before the start of the war and quantify the total number of buildings damaged within Gaza and its constituent governorates. The Humanitarian Open Street Map (HOTOSM) team \cite{hotosm} began an update to its layer on building-footprint data over Gaza following the onset of the war. The effort to complete the record included dozens of volunteers manually outlining building footprints from nadir viewing satellite optical imagery captured between 2019-2023 and sourced from Bing \cite{hotosm_gaza}. The updated records consist of 330,079 unique vector outlines representing building footprints distributed across the five governorates (Fig. \ref{fig:footprints}) and with an average building footprint area of 160 m$^2$. To our knowledge, these data constitute the most up-to-date manually-delineated building footprint dataset available as a pre-conflict reference over Gaza.

\subsubsection{UNOSAT damage locations}

\begin{figure}[H]
    \centering
    \centerline{\includegraphics[width=8cm]{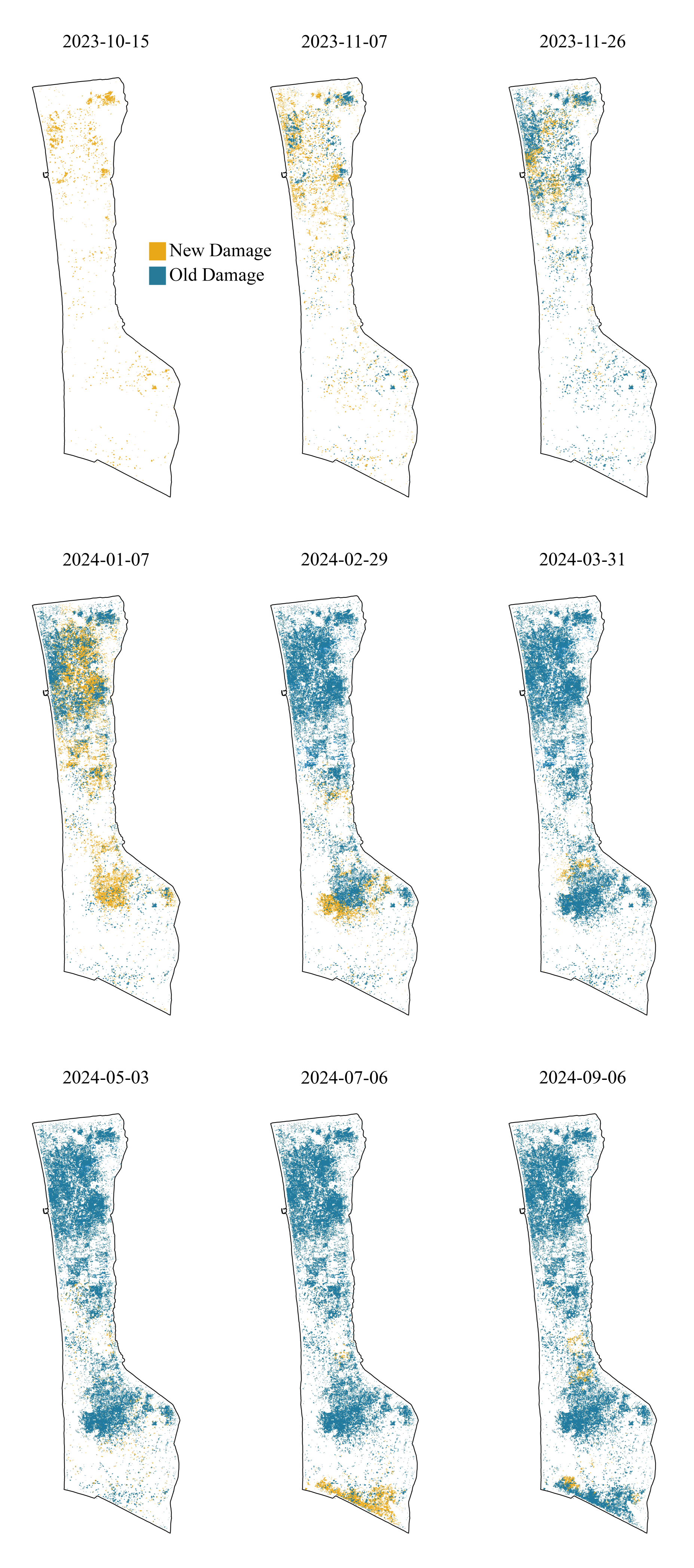}}
    \caption{UNOSAT data released during the study period aggregated to our 40m coherence pixel grid. All damage categories from UNOSAT are combined and grid cells with new damage are plotted in orange while grid cells with older damage are styled in blue.}
    \label{fig:unosat_damage}
\end{figure}

Damage locations visible in VHR satellite optical imagery are used to quantify agreement metrics against locations of radar-indicated damage. The United Nations Satellite Agency (UNOSAT) released nine sets of geospatial point vector data representing interpreted degrees of building damage. UNOSAT data are widely ingested by journalists and humanitarians in conflict settings, have been used as ground truth data in some studies (e.g., \cite{Bai2017Dec, Rathje2011Oct}), are not field validated, and are known to have errors of omission because not all forms of damage are visible to the naked eye in VHR optical data \cite{Rathje2011Oct}. Nonetheless, these data constitute perhaps the most complete and publicly-available representation of damage to buildings in Gaza across the study period. Point-level damage data mapped by UNOSAT include 163,778 locations of aggregate damage as of September 3-6, 2024 following analysis of 30 cm and 50 cm resolution optical data acquired by the Worldview-2 and Pleiades sensors (Fig. \ref{fig:unosat_damage}). UNOSAT generates these "comprehensive damage assessment" (CDA) data by labeling locations with four primary levels of interpreted building damage severity. These interpreted levels of damage severity have no basis in structural engineering for damage classification. Instead, these labels are intended to classify damage severity by categorizing the extent to which damage is visible in VHR imagery acquired overhead or slightly off-nadir.

UNOSAT reports damage with labels corresponding to interpeted damage severity. The most severe category of UNOSAT-interpreted damage is labeled as "destroyed" if the analyst interprets at least half of a structure to be collapsed \cite{iwg_sem_guidelines}. "Severe damage" is labeled if a roof has fully collapsed or parts of a structure are visibly collapsed. "Moderate damage" is assigned if an analyst observes damage to a rooftop and the emplacement of "large debris/rubble or sand deposits" around a building \cite{iwg_sem_guidelines}. A final category of "possible damage" is assigned if "small traces of debris/rubble or sand" is emplaced adjacent to a building. Possible damage is also labeled if damage interpretation is uncertain, or if a building is surrounded by damaged or destroyed buildings. For the first 6 UNOSAT CDA data releases, the "possible damage" category was omitted. Beginning in April of 2024, the "possible damage" category was added to the UNOSAT CDA data releases and marked a 39\% increase in the number of buildings labeled as damaged by UNOSAT between March and April. The possible damage category was then added retrospectively to all prior assessments, generating a record with consistent labels across the study period. UNOSAT generated nine CDA surveys in total across the study period, delineating 928,397 locations over time (Table \ref{tab:unosat_counts}). For three surveys, where VHR optical scenes analyzed by UNOSAT did not fully cover the Gaza Strip, images from two different dates with 1-3 days of lag time between acquisitions were combined to generate full coverage over Gaza. UNOSAT combined data from images acquired on January 6 and 7, 2024 to produce one Gaza-wide assessment in January 2024, similarly combined data from the end of March and early April for the April release, and finally combined data collected in early September for the final survey released during the study period (Table \ref{tab:unosat_counts}).

It is novel for UNOSAT to conduct repeat surveys across the same extent and at somewhat regular time intervals. UNOSAT released nine surveys during the study period with roughly one month of latency between image acquisition and publication of results, constituting a robust spatiotemporal set of point-level data on damage labels to compare to our automated CCD damage detections. Three of the surveys utilized VHR optical images acquired on different dates, but within a few days of each other (Supplementary table). We utilize these spatiotemporal UNOSAT data at each specific image date to report agreement in regions valid for CCD monitoring (Table \ref{tab:unosat_agreement_valid}) and for all UNOSAT locations including those outside of CCD-valid monitoring areas (Supplementary Table \ref{tab:unosat_agreement_agg}).

\begin{table}[H]
\begin{center}
\caption{Number of locations of damage or destruction reported by UNOSAT for each date of imagery analyzed.}
\begin{tabular}{lcccc}
\hline
 {} & Destroyed & Severe & Moderate & Possible Damage \\
 \hline
 \hline
2023-10-15 & 1922 & 2093 & 6286 & 3925 \\
2023-11-07 & 6806 & 5857 & 15311 & 11603 \\
2023-11-26 & 10109 & 7587 & 21381 & 15851 \\
2024-01-06 & 15090 & 9475 & 28187 & 16296 \\
2024-01-07 & 8377 & 3673 & 12420 & 8252 \\
2024-02-29 & 32433 & 15926 & 48006 & 27492 \\
2024-03-31 & 11485 & 4103 & 12637 & 6114 \\
2024-04-01 & 23633 & 12107 & 36772 & 21701 \\
2024-05-03 & 36935 & 16420 & 50005 & 38550 \\
2024-07-06 & 46532 & 18529 & 56221 & 40569 \\
2024-09-03 & 12693 & 3497 & 9800 & 6701 \\
2024-09-06 & 39871 & 15416 & 46910 & 34048 \\
\bottomrule
\end{tabular}
\label{tab:unosat_counts}
\end{center}
\end{table}

\subsection{Long temporal-arc coherent change detection (LT-CCD)}

Our method uses two single reference images and two dozen pre-war secondary Sentinel-1 images to construct two stacks of coherence estimates used for CCD at each time step. We select InSAR reference scenes acquired from similar orbital vantage points and at similar times in a seasonal cycle -- one during a pre-war period corresponding to each wartime acquisition. We design InSAR pairs to have matching distributions of temporal baselines across pre-war and wartime monitoring periods, and distill these coherence image stacks to stack-average and standard deviations for LT-CCD.

\begin{figure}[H]
    \centering
    \includegraphics[width=8cm]{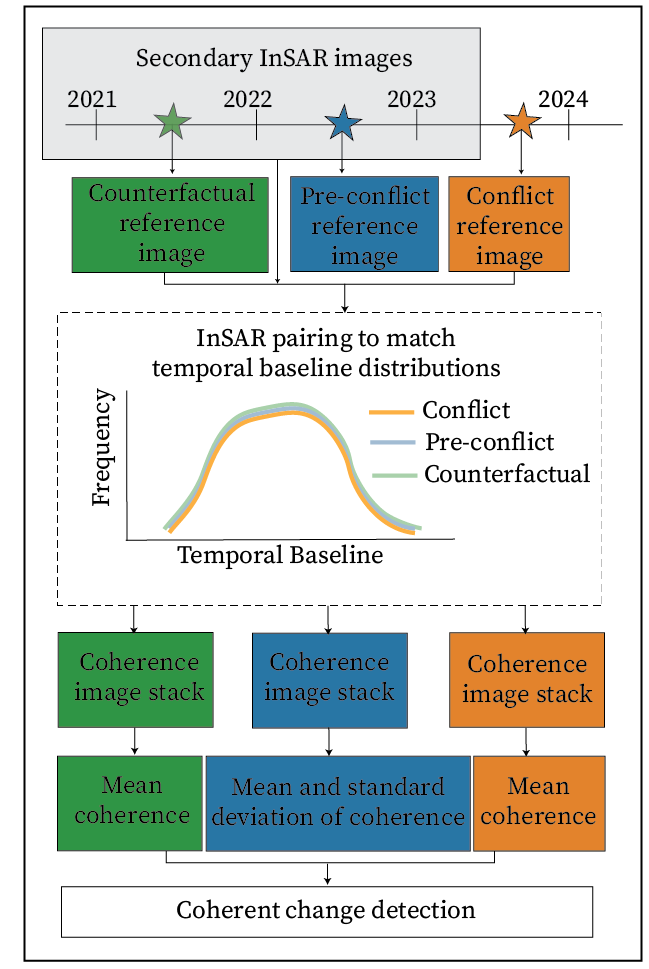}
    \caption{Pre-processing workflow to generate InSAR pairs and multi-temporal coherence estimates at one time step in CCD monitoring.}
    \label{fig:workflow}
\end{figure}

We draw from concepts in two modalities of radar interferometry for our approach to CCD. In persistent scatterer (PS) interferometry \cite{hooper2008multi}, long term coherence characteristics across multitemporal coherence image stacks are used to identify regions considered stable for monitoring sub-centimeter scale deformation with phase changes \cite{crosetto2016persistent}. Similarly, short baseline subsets (SBAS) interferometry \cite{Shanker2011Jan} is designed to minimize the effect of spatial (or perpendicular) and temporal baseline decorrelation using sets of InSAR pairs with minimum temporal and spatial baseline offsets for phase change applications. For each overpass we analyze during the conflict monitoring period, we assemble single-reference image stacks of InSAR pairs using each image acquired during the conflict and images acquired before the conflict, similar to single reference image InSAR stack formation for PS estimation. We then sample this stack of conflict period coherence images for the mean coherence at each pixel across the stack of data ($\overline{\gamma}_t$). Because of strong seasonal drivers related to coherence generally (e.g., \cite{kellndorfer2022global}), we select a single reference image acquired around the same time during the year prior and with the smallest perpendicular baseline available to construct a stack of pre-conflict InSAR pairs \ref{fig:workflow}). We select the acquisition with the shortest perpendicular baseline acquired acquired roughly one year before each conflict image acquisition. We intend to minimize differences in spatial baselines between pre- and conflict reference images while also using data acquired at a similar time of the year.

Building damage assessment with multi-temporal SAR using the average of coherence characteristics estimated across numerous combinations of InSAR pairs captures damage to buildings more accurately than the use of single coherence images for change detection \cite{Yang2024Jul, Monti-Guarnieri2018Jul}. We generate pixel-wise statistics on multi-temporal coherence characteristics as an approach intended to deal with exogenous drivers of coherence unrelated to building damage. This approach is informed by past work in space and time coherent change detection \cite{Monti-Guarnieri2018Jul} and has other more recent precedent \cite{Yang2024Jul}. For change detection at each time step in monitoring, we form one stack of up to 25 coherence images using a single reference image acquired during the conflict and reduce that stack of coherence images down to an average value. We assemble a stack of coherence images with a reference from a pre-conflict period and similarly reduce that stack to the mean and standard deviation of coherence at each pixel. These pixel-wise summaries of coherence characteristics during each epoch are the data that we use to conduct CCD.

\subsubsection{Conflict, pre-conflict, and counterfactual periods}

We define three epochs to detect changes during the conflict and quantify metrics for agreement \cite{SCHER2025100217}. A conflict monitoring period constrains the duration for damage monitoring (12 October 2023 - 31 October 2024). A pre-conflict baseline period establishes reference coherence characteristics to compare against coherence characteristics during the conflict. A final pre-conflict counterfactual period during a time without active fighting constrains false positive and true negative rates of detection of damage. Instead of quantifying false positives and true negatives using hits and misses between CCD and ancillary data on building damage during the conflict directly, because no data are field validated and a true accuracy assessment is not possible in this study, the counterfactual monitoring period serves to quantify rates of true negative (no damage classified) and false positive (damage falsely classified) during a time when active fighting and bombardment was not taking place. We will refer to the conflict, pre-conflict baseline, and counterfactual periods throughout the remainder of the text accordingly.

Sentinel-1A has three orbits with complete coverage over Gaza during the ongoing conflict: two ascending direction orbits (paths 160 and 87) and one descending orbit (path 94). In total, we consider 61 Sentinel-1 images with complete coverage over Gaza acquired during the conflict monitoring period. Each of these 61 images serves as a reference to form a total of 1,433 coherence images for the conflict monitoring period. We compare these conflict period coherence estimates to about as many coherence images assembled during the pre-conflict baseline period and, finally, quantify false positives and true negatives using coherence image stacks assembled during the pre-conflict counterfactual period (Table \ref{tab:image_counts}); where each counterfactual set of data corresponds to one of nine ancillary datasets on building damage to which we will assess for agreement against CCD.

\begin{table}[H]
\begin{center}
\caption{Count of unique Sentinel-1 image acquisitions during each monitoring epoch and the number of total coherence images generated for each epoch.}

\begin{tabular}{lrrr}
\toprule
 & Conflict & Pre-Conflict & Counterfactual \\
 \hline \hline
Sentinel-1 Acquisitions & 61 & 251 & 9 \\
Coherence Images & 1433 & 1414 & 442 \\
\bottomrule
\end{tabular}
\label{tab:image_counts}
\end{center}
\end{table}

\subsubsection{Matching distributions of temporal baselines}

\begin{figure}[H]
    \centering
    \includegraphics[width=12cm]{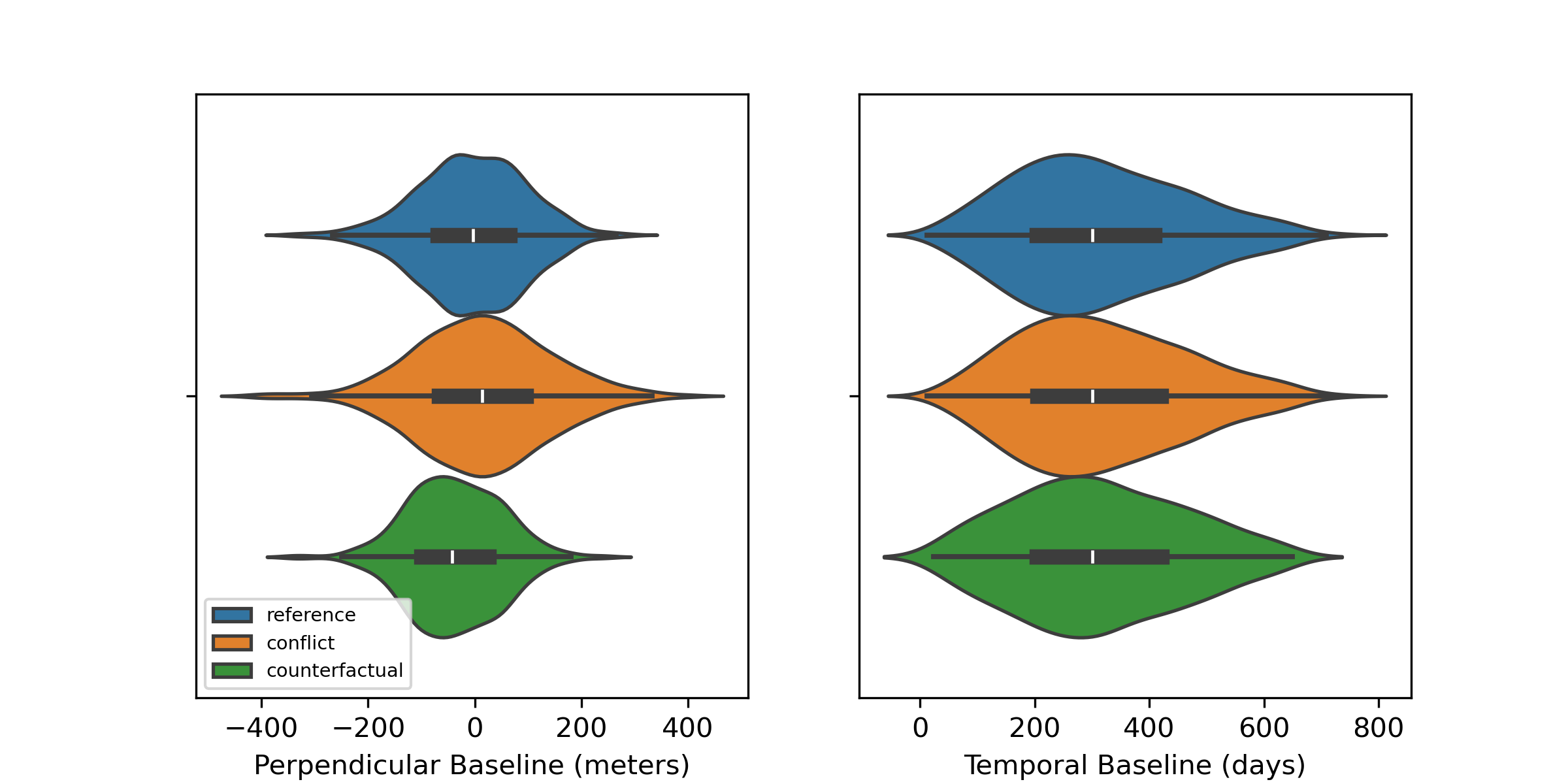}
    \caption{Distributions of perpendicular baselines and the absolute value of temporal baselines.}
    \label{fig:baselines}
\end{figure}

As the conflict progresses, the temporal offset between conflict acquisitions used as reference images for InSAR pairing and pre-conflict secondary images increases. This shifts the distribution of temporal baselines for each conflict monitoring period stack further from the pre-conflict stacks as the conflict proceeds, likely driving pronounced coherence decreases due to temporal decay alone \cite{zebker1992decorrelation, jung2016coherent}. To account for shifting distributions of temporal baselines over the study duration, we present an unconventional approach to InSAR pairing for CCD and assemble InSAR stacks using image pairs with matching distributions of the absolute value of temporal baselines across all epochs. This results in data across all three epochs with essentially identical distributions of temporal baselines and very similar distributions of perpendicular baselines (Fig. \ref{fig:baselines}).

\noindent
\subsubsection{Optimizing the number of InSAR pairs at each time step}

At each time step we assemble no more than 25 coherence images for each pre-, conflict, and counterfactual coherence image stack. We find that twenty five images is sufficient to converge on general tendencies without the need for additional data under consideration. In all cases, the average number of coherence images that we assemble for each epoch and at each time step is two dozen images. Failure of InSAR processing at HyP3 during some time steps results in small differences in the total number of coherence images formed at each time step but, in all cases, the number of coherence images is greater than or equal to 15 images, which we identify as a lower bound important to minimize residual noise due a reduced sample of coherence images. We seek to balance the volume of data under consideration with the speed and ability to converge on a similar result and find that a stack of images will converge toward a long-term average coherence value with at least 15 images and that the addition of images beyond 25 is not necessary given the additional computational costs.

\subsubsection{Damage classification}

For damage classification, we adapt a CCD approach for conflict damage mapping over long durations \cite{SCHER2025100217} and compare mean coherence estimates from each multi-temporal stack of coherence images during the conflict monitoring period ($\overline{\gamma}_{t}$) to the mean ($\overline{\gamma}_{pre}$) and standard deviation ($\sigma_{pre}$) of pre-conflict coherence at each time step using a fixed threshold and z-score metric. We classify potential damage if coherence decreases are below a fixed threshold ($k$) and if $\overline{\gamma}_{conflict}$ is two standard deviations below ($\overline{\gamma}_{pre}$) (Eq. \ref{eq:cd_threshold}-\ref{eq:damage_t}). Following initial detection of damage, we require that damage is detected again at least once during the following month of image acquisitions. This process is intended to remove false alarms from coherence decreases unrelated to signals of likely damage that do not recur across the record.

\begin{equation}
    {\Delta \gamma_{t}} = \overline{\gamma}_t -  \overline{\gamma}_{pre}
    \label{eq:cd_threshold}
\end{equation}

\begin{equation}
    {z_{t}} = \frac{\Delta \gamma_{t}}{\sigma_{\gamma_{pre}}}
    \label{eq:z_threshold}
\end{equation}

\begin{align}
        {Damage_{t}} = \begin{cases} {\Delta \gamma_{t}}  < k  & \text{CCD threshold met} \\
        {z_{t} < -2} & \text{z-score threshold met} \\
        \end{cases}
    \label{eq:damage_t}
\end{align}


\subsubsection{Classifying area valid for monitoring}

\begin{figure}[H]
    \centering
    \includegraphics[height=18cm]{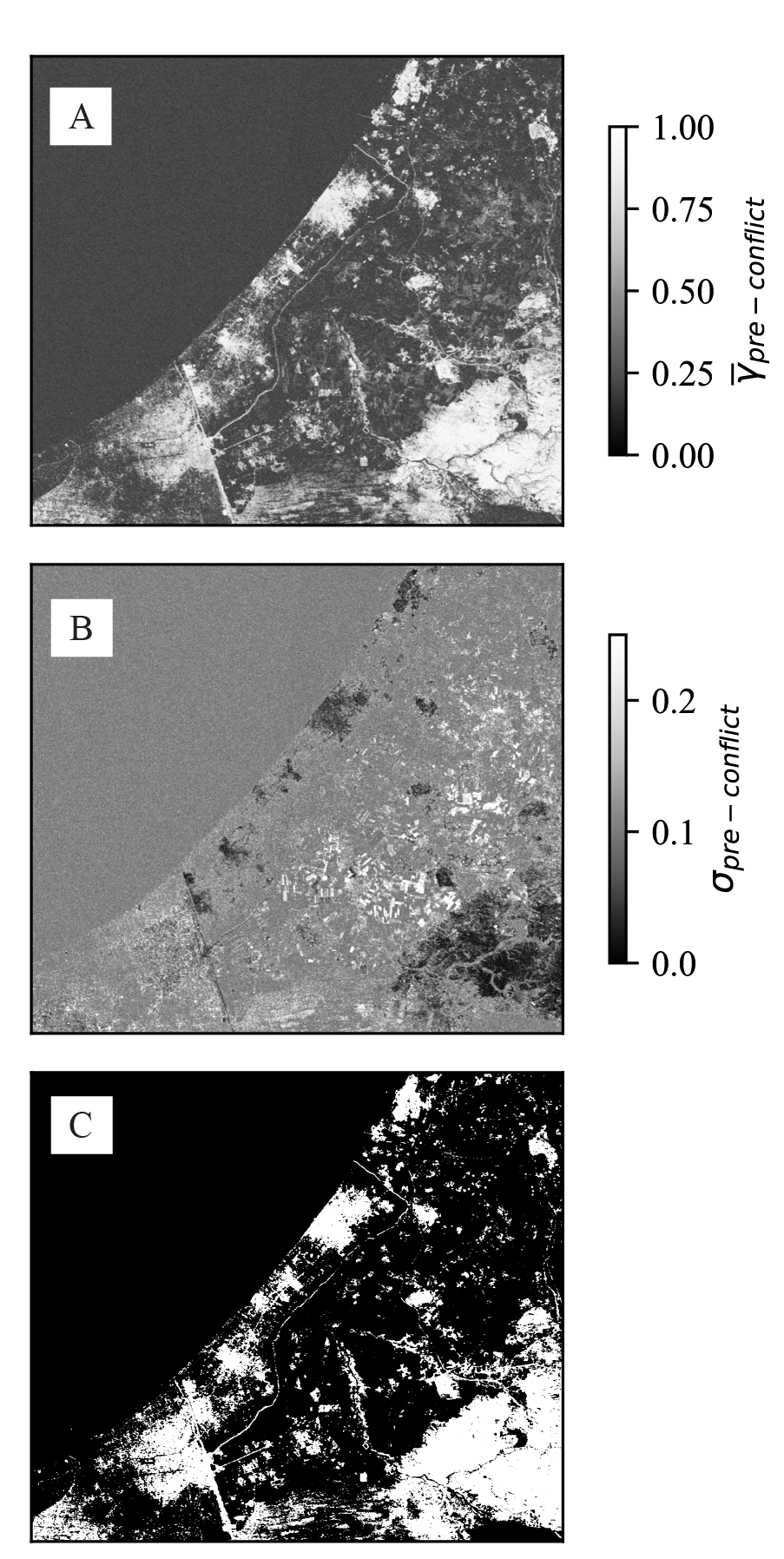}
    \caption{(A) Pixelwise mean pre-conflict coherence across a stack of images with a single reference image acquired on 17 December 2022. (B) Standard deviation of pre-conflict coherence across all coherence estimates generated using the same single reference image. (C) Areas valid for monitoring (white) and invalid (black) at this time step in monitoring.}
    \label{fig:valid_areas}
\end{figure}

We quantify which areas are valid for CCD monitoring using the z-score metric from Eq. \ref{eq:z_threshold} relative to the fixed threshold used for CCD ($k$). If the fixed threshold used for CCD has a pixelwise z-score two standard deviations below $\overline{\gamma}_{pre}$, that region is valid for damage monitoring. We illustrate the process for delineating image regions valid for monitoring in Figure \ref{fig:valid_areas}. This mask is important to quantify the portion of UNOSAT data that falls within areas valid for CCD monitoring and assess how well CCD monitors areas also monitored by UNOSAT. In total, 282,009 (85.58\%) of OSM building footprints fall within areas valid for monitoring during at least two CCD surveys, which would be the minimum needed to be classified as damage. Additionally, CCD is valid for monitoring 86.7\% (804,858) of UNOSAT locations over time.

\subsubsection{Accounting for number of buildings in damage-affected areas}

To tabulate the total number of damaged buildings, we calculate the fraction of each building footprint that is covered by a pixel labeled as damage and, if the fraction of the building footprint covering the pixel is greater than or equal to 99\% of the building footprint area, we mark that building as likely damaged or destroyed. To report on the total fraction of buildings damaged or destroyed in the Gaza Strip and constituent governorates, we calculate the ratio of building footprints labeled with CCD damage and divide by the total number of buildings mapped in each governorate. We then report on the fraction of buildings damaged in each governorate over time across the conflict monitoring period.

\subsubsection{Agreement with UNOSAT data}

We use CDA data from UNOSAT to quantify agreement metrics across conflict and counterfactual monitoring period epochs (Table \ref{table:agreement_metrics}). We quantify false positives and true negatives using coherence data from the counterfactual epoch where damage due to conflict did not take place. Because we do not attempt to predict the degree of damage severity at the level of an individual building and because past studies have found that VHR optical imagery can only reliably discriminate between full and partial building collapse \cite{Rathje2008May}, we combine all UNOSAT categories of damage into one category of damage or destruction, similar to other studies comparing detected damage with UNOSAT data \cite{Rathje2011Oct}. To quantify true positives and false negatives, we compare cumulative CCD-indicated damage at each time step following the release of each UNOSAT survey in Gaza. For each comparison with UNOSAT data, we select the nine CCD surveys immediately following the date of each image acquisition used for UNOSAT and accumulate the total damage indicated by CCD during both conflict and counterfactual periods up to the date of each UNOSAT survey; generating estimates of cumulative damage up through the point in time that each UNOSAT survey was conducted. To quantify false positives and true negatives, we aggregate all damage detected across the counterfactual period in an identical manner. We do not apply any criteria for persistence of CCD-damage signals in these agreement exercises because we seek to assess the performance of this approach for near-real time reporting on conflict impacts. The persistence criteria is intended as a quality-control mechanism for harmonized and retrospective damage data as monitoring persists across a conflict duration.

\begin{table}[H]
\begin{center}
\caption{Metrics of agreement for comparison of CCD damage data with UNOSAT-reported damage location.}
\renewcommand\arraystretch{2}
\begin{tabular}{ c c }
    \hline
    \hline
    True positive (TP) & InSAR-detected UNOSAT during the conflict period\\

    True negative (TN) & No damage classified during the counterfactual period \\

    False negative (FN) & InSAR-missed UNOSAT damage during the conflict period\\

    False positive (FP) & InSAR-detected UNOSAT during the counterfactual period\\

    Agreement & $\myfrac{TP+TN}{TP+TN+FP+FN}$\\

    True positive rate & $\myfrac{TP}{TP + FN}$ \\

    False positive rate & $\myfrac{FP}{TP + TN}$ \\

    CSI & $\myfrac{TP}{TP + FP + FN}$ \\

    F1 Score & $2 \times \frac{\text{TPR} \times (1 - \text{FPR})}{\text{TPR} + (1 - \text{FPR})}$\\

    \hline
\label{table:agreement_metrics}
\end{tabular}
\end{center}
\end{table}

\subsubsection{Assessing exogenous drivers of coherence decreases}
To gauge whether exogenous processes drive coherence decreases unrelated to building damage, we look at coherence characteristics in a region in Israel without reports of extensive damage. We use two metrics for comparison of coherence characteristics in this pseudo-stable region in Israel at each time step in monitoring. The first metric, a Hellinger distance between probability distributions of $\overline{\gamma}_{t}$ and $\overline{\gamma}_{pre}$ within a pseudo-stable sampling region, serves as an indicator of the similarity of the two distributions. A Hellinger distance has values ranging from 0 to 1. An identical pair of distributions will have a Hellinger distance of 0 and  probability distributions that are entirely dissimilar will have a Hellinger distance of 1. In addition to the Hellinger distance we look at the difference in average coherence between $\overline{\gamma}_{t}$ and $\overline{\gamma}_{pre}$ at each time step over the pseudo-stable area in Israel. Together, these metrics attribute metadata at each time step in monitoring indicative of the similarity of coherence characteristics across the pseudo-stable area in Israel. For each time step in monitoring, we compare each pre-conflict and during-conflict distribution of coherence values using each of these two metrics and summarize those comparisons across all time steps in the results below.


\section{Author contributions}

C.S. conceived of the approach for LT-CCD and refined the methods following extensive discussions and trials conducted in close collaboration J.V.D.H. C.S. engineered the data pipeline for scaling of InSAR processing to link ASF on-demand InSAR resources with cloud-based change detection using Google Earth Engine (GEE). Both C.S. and J.V.D.H. coordinated processing resources for uplift requests at ASF and GEE. J.V.D.H developed a code-base to post-process detected damage, aggregate damage detections by administrative units, and test for persistence of damage retrieval. C.S. authored the original manuscript and original figure artwork. J.V.D.H provided editorial review and direct suggestions to both the manuscript text and figure artwork. Both authors discussed the interpretation of the results and contributed to revising and finalizing the manuscript text.


\section{Supplementary materials}

\begin{table}[H]
\begin{center}
\caption{Agreement metrics between CCD and UNOSAT locations in areas valid for CCD monitoring at each time step where UNOSAT analyzed VHR satellite optical imagery for visible damage.}
\begin{tabular}{lcccccc}
\toprule
Image Date & Agreement & TPR (\%) & FPR (\%) &  F1 (\%) & CSI Score (\%) &  Total locations \\
\midrule
2023-10-15 & 77.98 & 56.25 & 0.29 & 71.87 & 56.09 & 11476 \\
2023-11-07 & 81.63 & 63.95 & 0.69 & 77.68 & 63.51 & 32479 \\
2023-11-26 & 85.91 & 72.58 & 0.75 & 83.75 & 72.04 & 46838 \\
2024-01-06 & 92.54 & 86.43 & 1.35 & 92.05 & 85.28 & 70340 \\
2024-01-07 & 84.83 & 70.61 & 0.94 & 82.32 & 69.95 & 27843 \\
2024-02-29 & 93.57 & 88.29 & 1.15 & 93.21 & 87.28 & 108248 \\
2024-03-31 & 92.33 & 85.43 & 0.78 & 91.76 & 84.77 & 30344 \\
2024-04-01 & 95.22 & 91.81 & 1.37 & 95.05 & 90.57 & 77596 \\
2024-05-03 & 94.45 & 90.17 & 1.26 & 94.21 & 89.05 & 119178 \\
2024-07-06 & 94.15 & 89.60 & 1.30 & 93.87 & 88.46 & 137287 \\
2024-09-03 & 91.53 & 83.77 & 0.71 & 90.82 & 83.18 & 29079 \\
2024-09-06 & 95.04 & 91.84 & 1.75 & 94.88 & 90.26 & 114150 \\
\bottomrule
\end{tabular}
\label{tab:unosat_agreement_valid}
\end{center}
\end{table}

\begin{table}[H]
\begin{center}
\caption{Agreement metrics between CCD and all UNOSAT locations at each time step.}
\begin{tabular}{lcccccc}
\toprule
Image Date & Agreement & TPR (\%) & FPR (\%) &  F1 (\%) & CSI Score (\%) &  Total locations \\
\midrule
2023-10-15 & 72.54 & 45.39 & 0.32 & 62.30 & 45.25 & 14226 \\
2023-11-07 & 75.96 & 52.53 & 0.60 & 68.60 & 52.21 & 39577 \\
2023-11-26 & 80.64 & 61.94 & 0.66 & 76.19 & 61.53 & 54937 \\
2024-01-06 & 86.47 & 74.15 & 1.21 & 84.57 & 73.27 & 82099 \\
2024-01-07 & 79.64 & 60.11 & 0.83 & 74.70 & 59.62 & 32722 \\
2024-02-29 & 88.09 & 77.24 & 1.06 & 86.64 & 76.43 & 123859 \\
2024-03-31 & 88.49 & 77.71 & 0.72 & 87.10 & 77.15 & 33378 \\
2024-04-01 & 88.87 & 78.97 & 1.24 & 87.64 & 78.00 & 90333 \\
2024-05-03 & 88.77 & 78.67 & 1.13 & 87.50 & 77.79 & 136762 \\
2024-07-06 & 88.71 & 78.58 & 1.17 & 87.44 & 77.68 & 156707 \\
2024-09-03 & 88.20 & 77.07 & 0.67 & 86.73 & 76.56 & 31611 \\
2024-09-06 & 88.93 & 79.41 & 1.55 & 87.77 & 78.20 & 132186 \\
\bottomrule
\end{tabular}
\label{tab:unosat_agreement_agg}
\end{center}
\end{table}

\bibliography{bib}

\end{document}